\documentclass[10pt,journal,compsoc]{IEEEtran}
\usepackage{booktabs}
\usepackage{graphicx}
\usepackage{graphics}
\usepackage{placeins}
\usepackage{float}
\usepackage{subfigure}
\usepackage{amsmath}
\usepackage{url}

\usepackage[table]{xcolor}
\usepackage{collcell}
\usepackage{hhline}
\usepackage{pgf}
\usepackage{multirow}
\usepackage{tikz}
\usetikzlibrary{calc}
\usepackage {color, graphicx}
\usepackage {subfigure}
\usetikzlibrary{decorations.pathmorphing}
\usetikzlibrary {patterns,shapes,shadows,decorations.pathreplacing}
\usetikzlibrary{positioning}
\usepackage{amssymb}

\ifCLASSOPTIONcompsoc
  \usepackage[nocompress]{cite}
\else
  \usepackage{cite}
\fi
\ifCLASSINFOpdf
\else
\fi
\begin{document}
\title{Measuring Human and Economic Activity from Satellite Imagery to Support City-Scale Decision-Making during COVID-19 Pandemic}

\author{ Rodrigo~Minetto,~\IEEEmembership{Member,~IEEE,}
         Maur\'icio~Pamplona~Segundo,~\IEEEmembership{Member,~IEEE,} \\
         Gilbert Rotich and 
         Sudeep~Sarkar,~\IEEEmembership{Fellow,~IEEE}% <-this % stops a space
\IEEEcompsocitemizethanks{
\IEEEcompsocthanksitem R. Minetto is with Universidade Tecnol\'{o}gica Federal do Paran\'{a} (UTFPR), Brazil. E-mail: rodrigo.minetto@gmail.com
\IEEEcompsocthanksitem M. Pamplona Segundo, G. Rotich and S. Sarkar are with Department of Computer Science and Engineering, University of South Florida (USF), Tampa, FL, USA. E-mail: \{mauriciop,grotich,sarkar\}@usf.edu
%\IEEEcompsocthanksitem G. Rotich is with Department of Computer Science and Engineering, University of South Florida (USF), Tampa, FL, USA. E-mail: grotich@mail.usf.edu.
%\IEEEcompsocthanksitem S. Sarkar is with Department of Computer Science and Engineering, University of South Florida (USF), Tampa, FL, USA. E-mail: sarkar@usf.edu
}% <-this % stops a space
%\thanks{Manuscript received December ??, 20??; revised August ??, 20??.}}
}

\markboth{The final version of this work is available on IEEE: https://doi.org/10.1109/TBDATA.2020.3032839}%
{Minetto \MakeLowercase{\textit{et al.}}: Measuring Human and Economic Activity from Satellite Imagery to Support City-Scale Decision-Making during COVID-19 Pandemic}
\maketitle

%\copyrightnotice

% is not selected <OR> if conference mode is selected - because compsoc
% conference papers position the abstract like regular (non-compsoc)
%\IEEEdisplaynontitleabstractindextext

% \fi
%\IEEEpeerreviewmaketitle

%\IEEEtitleabstractindextext{%
The COVID-19 outbreak forced governments worldwide to impose lockdowns and quarantines to prevent virus transmission. As a consequence, there are disruptions in human and economic activities all over the globe. The recovery process is also expected to be rough. Economic activities impact social behaviors, which leave signatures in satellite images that can be automatically detected and classified.  Satellite imagery can support the decision-making of analysts and policymakers by providing a different kind of visibility into the unfolding economic changes.  In this work, we use a deep learning approach that combines strategic location sampling and an ensemble of lightweight convolutional neural networks (CNNs) to recognize specific elements in satellite images that could be used to compute economic indicators based on it, automatically. This CNN ensemble framework ranked third place in the US Department of Defense xView challenge, the most advanced benchmark for object detection in satellite images. We show the potential of our framework for temporal analysis using the US IARPA Function Map of the World (fMoW) dataset. We also show results on real examples of different sites before and after the COVID-19 outbreak to illustrate different measurable indicators. Our code and annotated high-resolution aerial scenes before and after the  outbreak are available on GitHub\footnote{https://github.com/maups/covid19-satellite-analysis}.

%}

%%%%%%%%%%%%%%%%%%%%%%%%%%%%%%%%%%%%%%%%%%%%%%%%%%%%%%%%%%%%%%%%%%%%%%%%%%%%%%%%
\section{INTRODUCTION}\label{sec:introduction}
The COVID-19 outbreak is changing the world as never seen before. The lockdowns and quarantines implemented worldwide can be noticed even from space. Spatial agencies such as the US National Aeronautics and Space Administration (NASA) and the European Space Agency (ESA) observed a significant decrease in nitrogen dioxide emissions over major metropolitan areas around the world as a consequence of the economic slowdown. However, the potential use of remote sensing data goes far beyond. As an example, the European Union Commission requested the sharing of any satellite imagery related to the pandemic for research purposes\footnote{https://www.euspaceimaging.com/eu-commission-asks-eo-community-for-help-with-covid-19/}. Such images will support decisions concerning:
%\begin{itemize}
(1) traffic issues, to ensure citizens' mobility but at the same time to avoid traffic jams that block the exchange of essential supplies;
(2) medical infrastructure, to have knowledge about any temporary medical facility construction around Europe and to gain awareness on the impacts and actions taken in the face of the outbreak;
(3) facilities activity, to safely and economically maximize resources; and
(4) social distancing, to appraise if people are following orders during a quarantine.
%\end{itemize}

\begin{figure}[!htb]
    \centering
    \subfigure[Airport before and after the COVID-19]{
      \label{fig:covid-a}
      \hspace{-6pt}
      \includegraphics[width=4.22cm,height=2.7cm]{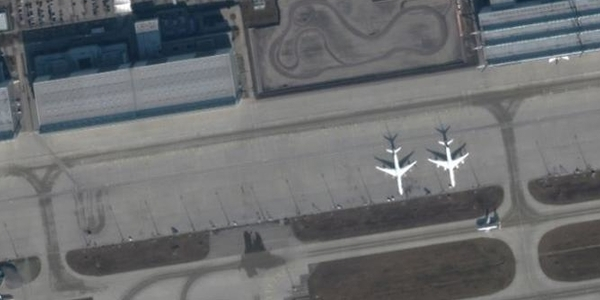}
      \includegraphics[width=4.22cm,height=2.7cm]{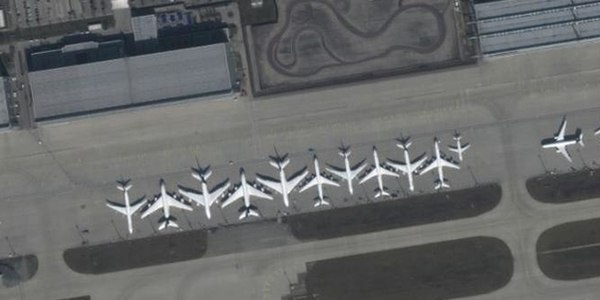}
    }
    \subfigure[Car rental parking lot before and after the COVID-19]{
      \label{fig:covid-b}
      \hspace{-6pt}
      \includegraphics[width=4.22cm,height=2.7cm]{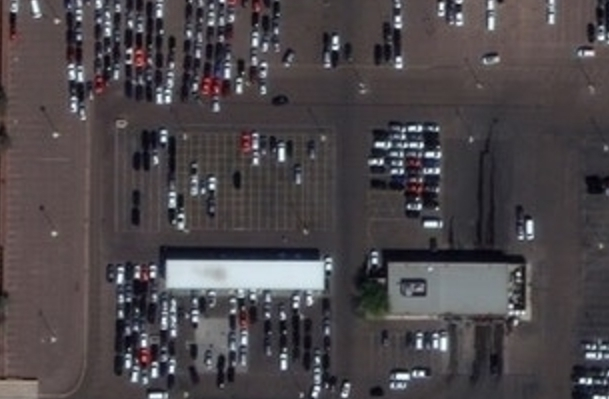}
      \includegraphics[width=4.22cm,height=2.7cm]{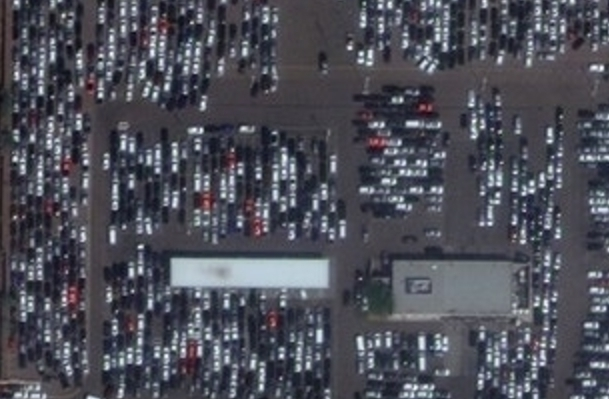}
    }
   \caption{COVID-19 impacts on human and economic activities. \textbf{Photo credit}: \textsc{Satellite image 2020 Maxar Technologies}.}
   \label{fig:covid}
\end{figure}

High-resolution imagery, as provided by sophisticated satellites like WorldView-3~\cite{ASADZADEH2016162} that collect panchromatic images daily with a ground sample distance (GSD) of 0.3 meters around the globe, can be a valuable asset to estimate the impacts of COVID-19 in society. Figure~\ref{fig:covid} presents two scenarios in which the analysis of strategic sites can provide critical indicators of human and economic activities over time. Figure~\ref{fig:covid-a} shows parked aircraft before and after the COVID-19 outbreak, while Figure~\ref{fig:covid-b} shows a car rental parking lot within a similar time frame. These examples illustrate the decrease in traveling caused by this pandemic and its consequential impact on aviation and car rental businesses. Other examples include obtaining information on traffic or distancing issues through detecting and classifying vehicles; keeping track of new medical infrastructure being built by identifying construction elements such as bulldozers, excavators, trucks, and tents; and measuring economic activity by detecting commercial transports such as planes, ships, and locomotives.

Although many indicators computed through remote sensing are also measurable from other data sources, the former is advantageous for its versatility. Monitoring systems based on satellite images can support new indicators and new areas of interest with little effort, as all of them share the same database and input format. Besides, scaling these systems up to a global level is a matter of satellite coverage and computational power, eliminating complications associated with data collection from heterogeneous sources at this scale. These characteristics favor the adaptation and application of such systems when a fast response is critical.

Nevertheless, to unleash the full potential use of satellite images, we need automated AI-based computer algorithms to extract these kinds of information from them, without requiring extensive manual labor, so local decision-makers all over the world can use them without time lag. As our main contribution, we present a framework that recognizes specific elements in strategic locations to compute such indicators automatically. As part of this work, we describe an ensemble of convolutional neural networks (CNN) for simultaneous detection and classification of objects in high-resolution aerial images. This approach is state of the art and ranked third place in the US Department of Defense conducted xView challenge~\cite{xview2018}. The xView dataset is very relevant to the COVID-19 problem because it used WorldView-3 as a source for more than 1,100 high-resolution images spanning about 800,000 aerial objects around the world, and covering a total area of 1,400 square kilometers. The organizers provided annotations for 60 classes of objects, with many of them being particularly relevant to the task of this work. We employ a combination of strategic location sampling and a lightweight CNN architecture to perform satellite image processing and analysis within an acceptable time frame. With that, we hope to support regular economic assessment and decision making processes. Furthermore, we manually annotated nearly 16,500 objects from high-resolution aerial scenes before and after the COVID-19 outbreak. We made them publicly available in our github repository hoping that they will be useful to other researchers addressing in the same problem.

We present our framework within the context of stay-at-home order enforcement (Section~\ref{sec:framework}) and discuss later how to adapt it to other scenarios (Section~\ref{sec:casestudies}). In our experiments, we first evaluate the detection performance on the xView dataset (Section~\ref{sec:exp_detector}) and then show its potential for temporal analysis using the US IARPA Functional Map of the World (fMoW) dataset~\cite{christie2018functional} (Section~\ref{sec:exp_temporal}). Finally, we show our framework in action on real examples of world scenes before and after the COVID-19 outbreak (Section~\ref{sec:casestudies}). 

\begin{figure*}[!htb]
\centering

\begin{tikzpicture}

\node[anchor=north west, inner sep=0pt, opacity=1.0] (img1) at (0.0,0.0) { \includegraphics[height=3.8cm]{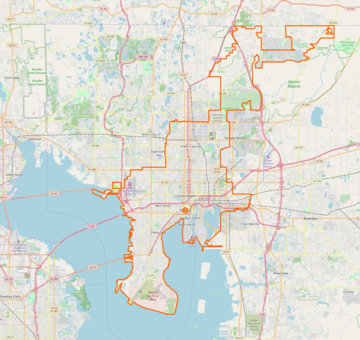} };
\path[->](2.0,0.25) node[black,very thick,inner sep=0pt] {\small (a) Area of interest};

\node[anchor=north west, inner sep=0pt, opacity=1.0] (img2) at (4.5,0.0) { \includegraphics[height=3.8cm]{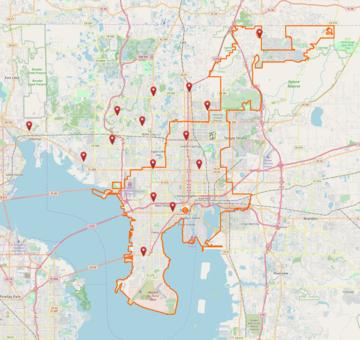} };
\path[->](6.45,0.25) node[black,very thick,inner sep=0pt] {\small (b) Location sampling};

\draw (img1) edge[->] (img2);

\node[anchor=north west, inner sep=0pt, opacity=1.0] (img3) at (9.0,0.0) { \includegraphics[height=3.8cm]{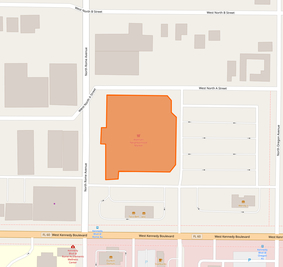} };
\path[->](11.0,0.25) node[black,very thick,inner sep=0pt] {\small (c) Location boundaries};

\draw (img2) edge[->] (img3);

\node[anchor=north west, inner sep=0pt, opacity=1.0] (img4) at (13.5,0.0) { \includegraphics[height=3.8cm]{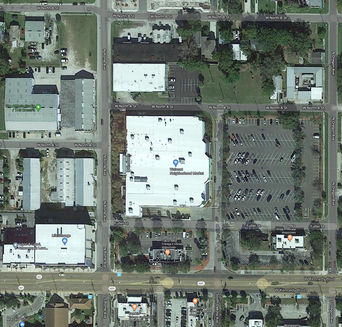} };
\path[->](15.40,0.25) node[black,very thick,inner sep=0pt] {\small (d) ROI extraction};

\draw (img3) edge[->] (img4);

\node[anchor=north west, inner sep=0pt,opacity=1.0] (img5) at (13.5,-4.5) { \includegraphics[height=3.8cm]{FIG-map-roi-sat.png} };
\begin{scope}[xshift=13.5cm,yshift=-4.5cm]
  \fill[white,opacity=0.5] (0.0,0.0) rectangle (4.0,-1.3);
  \fill[white,opacity=0.5] (0.0,-1.3) rectangle (2.6,-3.0);
  \fill[white,opacity=0.5] (0.0,-2.8) rectangle (4.0,-3.80);

  \draw[step=1.4cm,black=0.9,thick,dashed,xshift=0.0cm,yshift=-2.8cm] (0.0,0.0) grid (2.8,2.8);
  \draw[step=1.4cm,black=0.9,thick,dashed,xshift=2.6cm,yshift=-2.8cm] (0.0,0.0) grid (1.4,2.8);
  \draw[step=1.4cm,black=0.9,thick,dashed,xshift=0.0cm,yshift=-3.8cm] (0.0,0.0) grid (2.8,1.4);
  \draw[step=1.4cm,black=0.9,thick,dashed,xshift=2.6cm,yshift=-3.8cm] (0.0,0.0) grid (1.4,1.4);
\end{scope}
\path[->](15.30,-8.60) node[black,very thick,inner sep=0pt] {\small (e) Resize \& split};

\draw (img4) -- (17.8,-1.83) -- (17.8, -6.25) edge[->] (img5);

\node[anchor=north west, inner sep=0pt, opacity=1.0] (img6c) at (9.35,-4.85) { \includegraphics[angle=90,origin=c,height=3.0cm]{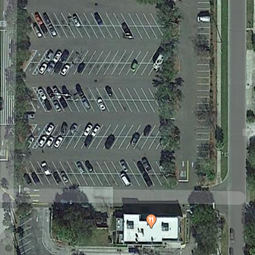} };
\draw[black=0.9,very thick] (9.35,-4.85) rectangle (12.35,-7.85);
\node[anchor=north west, inner sep=0pt, opacity=1.0] (img6b) at (9.175,-4.675) { \includegraphics[angle=180,origin=c,height=3.0cm]{FIG-map-roi-sat-split.png} };
\draw[black=0.9,very thick] (9.175,-4.675) rectangle (12.175,-7.675);
\node[anchor=north west, inner sep=0pt, opacity=1.0] (img6a) at (9.0,-4.5) { \includegraphics[height=3.0cm]{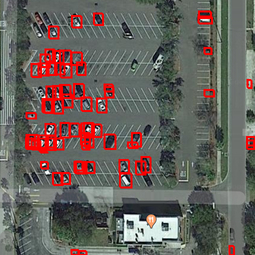} };
\draw[black=0.9,very thick] (9.0,-4.5) rectangle (12.0,-7.5);
\path[->](10.75,-8.60) node[black,very thick,inner sep=0pt] {\small (f) Vehicle detection};

\draw (img5) edge[->] (img6c);

\node[anchor=north west, inner sep=0pt, opacity=1.0] (img7) at (4.5,-4.5) { \includegraphics[height=3.8cm]{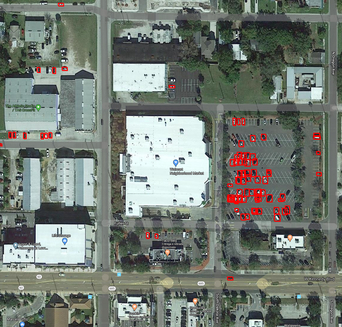} };
\path[->](6.29,-8.60) node[black,very thick,inner sep=0pt] {\small (g) Filter \& merge};

\draw (9.0,-6.4) edge[->] (img7);

\node[anchor=north west, inner sep=0pt, opacity=1.0] (img8) at (0.0,-4.5) { \includegraphics[height=3.8cm]{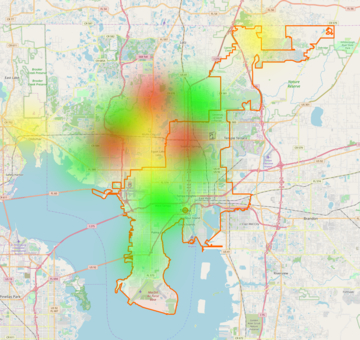} };
\path[->](1.95,-8.60) node[black,very thick,inner sep=0pt] {\small (h) Compliance evaluation};

\draw (img7) edge[->] (img8);

\end{tikzpicture}

\caption{We present a workflow to analyze the pattern of vehicles over time to monitor compliance with stay-at-home order. Similar workflows are possible for other aspects of the economy, such as supply-chain disruptions. Other than the first step, the rest is fully automatic.  (a) A human demarcates an area of interest.  (b) The algorithm samples strategic locations. (c) The algorithm then looks for their boundaries in open knowledge sources to delimit regions to arrive at (d) regions of interest (ROI) in satellite images.  (e) Each ROI is then automatically resized and split into several small parts to be (f) processed by a vehicle detector. (g) And the results are filtered and merged into a single outcome. (h) Finally, and this part is still conceptual, an algorithm will analyze the history of vehicle occupancy in each sampled location to help identify non-compliant zones. Maps, landmarks, and boundaries were obtained from OpenStreetMap\textsuperscript{*}. Satellite images were obtained from Google Maps\textsuperscript{**}.}
\vspace{0.1cm}
\scriptsize*~\textcopyright~OpenStreetMap contributors\hfill**~Image~\textcopyright~2020 Google, Maxar Technologies\hfill~\hfill~\hfill~\hfill~\hfill~
\label{fig:framework}
\end{figure*} 

\section{Related Work}
The use of satellite imagery is of paramount importance to support the management of natural disasters, humanitarian assistance, and environmental conservation policies. In recent years, the unprecedented amount of data captured by sophisticated satellites has profoundly impacted the information quality and the demand for techniques to extract knowledge from it.

The rise in the number of large-scale challenge datasets that has recently become available to foster breakthroughs in this field has been remarkable. The SpaceNet challenge~\cite{DBLP:journals/corr/abs-1807-01232} focused on the automated building footprint extraction and road network detection. The organizers used the WorldView-2 satellite to collect high-resolution images to cover more than 683,000 buildings and 8,676 road stretches of five metropolitan areas. As observed by them, these mappings are of particular interest in natural disasters. By using a satellite with a daily revisit time, it would be possible to quickly identify damaged buildings and blocked/destroyed roads and prepare the logistics for humanitarian assistance accordingly. This topic was later embraced by the xView2 challenge~\cite{CreatingxBD}, which focused on assessing building damage after a natural disaster. The organizers released pre- and post-disaster images from 850,000 buildings around the world, depicting the effects of earthquakes, tsunamis, floodings, volcanic eruptions, wildfires, tornados, and hurricanes. The US IARPA Functional Map of the World (fMoW) challenge~\cite{christie2018functional} encouraged the design of automated solutions for land use classification in satellite images. Its dataset comprised more than one million excerpts of multispectral images from 63 categories, including satellite metadata and temporal views. Among these categories are hospitals, educational institutions, airports, prisons, parks, electric and fire substations, places that are worth monitoring during a pandemic.

Remote sensing also plays a vital role in the study of human diseases, with many works stipulating associations between terrain characteristics and disease incidence. Rogers~\cite{ROGERS1991} observed a correlation between African trypanosomiases causing sleeping sickness and indices of temperature, rainfall, and vegetation obtained from satellite imagery. Rogers~\emph{et~al.}~\cite{ROGERS2002} later perceived that sensing seasonal climate could help to predict mosquito vectors that are responsible for malaria transmission. Dister~\emph{et~al.}~\cite{Dister1997} investigated the relationship between Lyme disease and measurements of vegetation structure, wetness, and abundance. Cyranoski~\cite{Cyranoski2009} mapped wetlands to study the spreading of avian influenza.  Ford~\emph{et~al.}~\cite{Ford2009} showed how sea surface temperature, sea surface height, and chlorophyll A levels can be used to predict outbreaks of cholera. Garni~\emph{et~al.}~\cite{GARNI2014} used land cover and topography information to map the risk of occurrence of cutaneous leishmaniasis.

In the field of economics, satellite images have helped to estimate different indicators. There are plenty of works based on satellite-recorded nighttime lights, as they provide a reasonable valuation of economic activity. Regression of gross domestic product (GDP)~\cite{Ghosh2010, Hu2019}, poverty levels~\cite{Elvidge2009, Noor2008} and development indices~\cite{Ghosh2010a} based on this information were deemed plausible in the literature. Recently, the analysis of high-resolution daytime satellite images improved such measurements~\cite{Jean2016, Suraj2017} thanks to the advances brought by deep learning~\cite{Lecun2015}. Other relevant efforts in this field include estimating asset wealth across thousands of African villages from publicly-available multispectral satellite imagery~\cite{Yeh2020} and predicting key food security metrics such as z-scores of stunting or wasting~\cite{ganguli2019predicting}.

The bio-inspired CNN~\cite{Fukushima1980}, a popular deep learning choice nowadays, is composed of multiple layers of artificial neurons and is used to learn representations with various levels of abstraction. Its ability to discover intricate patterns in massive data~\cite{Krizhevsky2012} has made it a perfect tool for remote sensing. It currently supports a myriad of applications in the literature, such as semantic segmentation~\cite{Kampffmeyer_2016_CVPR_Workshops}, target localization~\cite{Guirado2019}, region classification~\cite{8698456, 8713925}, image retrieval~\cite{8880494}, super-resolution~\cite{7937881}, regression models for environmental knowledge extraction~\cite{Kulp2019}, understanding of temporal and spatial variations~\cite{8316243}, study of semantic relationships between aerial targets~\cite{8707405}, 3D reconstruction~\cite{Leotta_2019_CVPR_Workshops}, and hyperspectral image generation~\cite{8241773}. As detailed in the next section, we also use deep learning as a tool to extract knowledge from satellite imagery. The main difference to other works is that we do not regress indicators directly from the images, but from information obtained from them, such as the number of vehicles, trucks, buildings, and so on.  This strategy allows us to create indicators that are informative, understandable, and supportive in decision-making.

%%%%%%%%%%%%%%%%%%%%%%%%%%%%%%%%%%%%%%%%%%%%%%%%%%%%%%%%%%%%%%%%%%%%%%%%%%%%%%%%
\section{Proposed Framework}\label{sec:framework}
The idea of analyzing the flow of vehicles under a stay-at-home order in a large area, such as a city or a county, using satellite images is hard to execute due to the vast amount of data to be processed. Thus, data sampling is necessary to reduce the computational cost so that it is possible to generate content to aid hazard assessment and decision making by authorities within an acceptable time frame. The sampling strategy, however, has to take the relevance of the selected regions to the problem into account. This because traditional sampling methods, such as random and grid sampling, tend to pick too many meaningless regions, which could lead authorities to incorrect assumptions. In a stay-at-home scenario to avoid disease proliferation, like the ones occurring due to the recent COVID-19 outbreak, the surrounding of places that gather crowds, like airports, schools, hospitals, churches, malls, and supermarkets, should be prioritized over underpopulated areas.

In this work, we present a complete framework to map increases and decreases in the flow of vehicles over an area of interest by combining open knowledge sources, such as OpenStreetMaps\footnote{https://www.openstreetmap.org}, satellite images, and a machine learning-based vehicle detector. Figure~\ref{fig:framework} illustrates the sequence of stages that compose our proposed framework. These stages are detailed in the following sections.

\subsection{Location sampling and region of interest extraction}
The first stage in our frameworks consists of a strategic sampling of locations within an area of interest. To do so, first, we need to define what is a strategic location. In this work, it can be any place with a high circulation of people that may contribute to the proliferation of pathogens. More specifically, we look for items with the following tags in the OpenStreetMap database: \emph{`shop=supermarket'}, \emph{`aeroway=aerodrome'}, \emph{`amenity=hospital'}, \emph{`amenity=university'}, \emph{`amenity=school'}, \emph{`shop=mall'}, and \emph{`amenity=place\_of\_worship'}. This tag list can be easily extended if necessary, or even redesigned for other applications.

Among the recovered items within the area of interest (see Figure~\ref{fig:framework}(b)), we select the ones that contain annotations for the boundary contour (see Figure~\ref{fig:framework}(c)) in the form of a sequence of latitude and longitude coordinates. We find the smallest enclosing bounding box for the coordinates of each item, which is then expanded $m$ meters in all directions to delimit the item's region of interest (ROI) in a satellite image (see Figure~\ref{fig:framework}(d)). Smarter ROI extraction strategies can be used, such as parking lot detection near strategic locations, upon the availability of reliable techniques and resources to support them.

\subsection{Vehicle detection}\label{sec:car-detection}
The input for this stage is a group of ROI images extracted in the previous stage, and the output for the $i$-th ROI is a set of $n_i$ detected regions $\mathcal{R}^i = \{r^i_1, r^i_2, \dots, r^i_{n_i}\}$. Each region $r_j$ is defined by an axis-aligned rectangular box $b(r_j) = (x_1, y_1, x_2, y_2)$ where $(x_1,y_1)$ and $(x_2,y_2)$ represent the upper left and bottom right corners, respectively. A score $w(r_j)$ expresses the confidence of the detection within the interval $[0,1]$.

In this work, we create an ensemble of Single Shot Multibox Detectors (SSD)~\cite{liu2016ssd} for vehicle detection. To do so, we combine two models released as baselines for the xView dataset~\cite{xview2018}, \emph{Vanilla} and \emph{Multires}, by using different parameters for image resizing, splitting, and output merging.

Even though ROIs are small parts of satellite images, they may still be too large for carrying out vehicle detection directly. Both baseline models receive $300\times300$ images as input. So we split our ROI images into blocks of $300\times300$ pixels without overlap whenever possible for the \emph{Vanilla} model, as illustrated in Figure~\ref{fig:framework}(e), or with an overlap of $100$ pixels for the \emph{Multires} model. Adding an overlap helps to detect vehicles that lay at the edge of two or more adjacent blocks. Besides, a second copy of the \emph{Vanilla} model uses ROI images scaled by a factor of $1.3$ to increase the detection accuracy of smaller vehicles. Table~\ref{tab:parameters_table2} summarizes this arrangement.

\begin{figure}[!htb]
\centering
\begin{tikzpicture}

\draw(0.0,-1.0) node[inner sep=0pt, opacity=0.9] (img1) { \includegraphics[width=2.8cm,height=2.8cm]{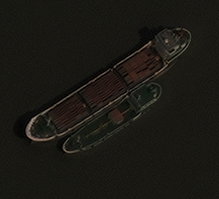} };
\draw[white=0.9,thick] (-1.1,0.1) rectangle (+1.1,-1.6);
\draw[white=0.9,thick] (-0.6,-0.7) rectangle (+0.7,-1.8);

\draw(2.9,-1.0) node[inner sep=0pt, opacity=0.9] (img1) { \includegraphics[width=2.8cm,height=2.8cm]{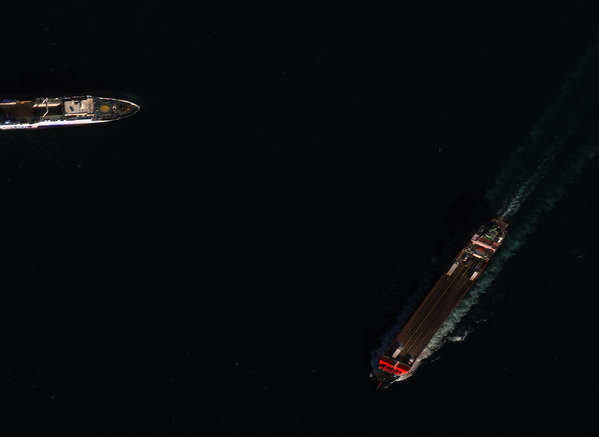} };
\begin{scope}[xshift=2.9cm,yshift=-4.1cm, scale=0.7]
  %-----------Box1
  \draw[white=0.9,thick] (-1.95,5.65) rectangle (-1.0,5.15);
  \draw[white,fill=white] (-1.95,5.65) circle (.3ex);
  \draw[white,fill=white] (-1.0,5.15) circle (.3ex);
  \path[->](-1.45,5.85) node[white,very thick, inner sep=1pt]  {\scriptsize $\bar{r}_2$};
 
  %-----------Box2
  \draw[white=0.9,thick] (+0.5,4.0) rectangle (+1.5,2.9);
  \draw[white,fill=white] (+0.5,4.0) circle (.3ex);
  \draw[white,fill=white] (+1.5,2.9) circle (.3ex);
  \path[->](+0.4,3.6) node[white,very thick, inner sep=1pt]  {\scriptsize $\bar{r}_3 (w = 0.8)$};
   
  %-----------Box3
  \draw[white=0.9,thick] (+0.4,4.9) rectangle (+1.8,2.82);
  \draw[white,fill=white] (+0.4,4.9) circle (.3ex);
  \draw[white,fill=white] (+1.8,2.82) circle (.3ex);
  \path[->](+1.0,5.1) node[white,very thick, inner sep=1pt]  {\scriptsize $\bar{r}_1 (w = 0.7)$};
  
  \path[->](-0.7,2.1) node[black,very thick,fill=white,fill opacity=0.9, inner sep=-2pt]  {\tiny $\bar{\mathcal{R}} = \{\bar{r}_1, \bar{r}_2, \bar{r}_3\}$};
  \path[->](3.9,2.1) node[black,very thick,fill=white,fill opacity=0.9, inner sep=1pt]  {\tiny $\textrm{NMS}(\bar{\mathcal{R}}_1 = \{\bar{r}_2\}) = r_1$};
  \path[->](4.15,1.7) node[black,very thick,fill=white,fill opacity=0.9, inner sep=1pt]  {\tiny $\textrm{NMS}(\bar{\mathcal{R}}_2 = \{\bar{r}_1, \bar{r}_3\}) = r_2$};
  \path[->](3.23,1.3) node[black,very thick,fill=white,fill opacity=0.9, inner sep=1pt]  {\tiny $\mathcal{R} = \{r_1, r_2\}$}; 
\end{scope}

\draw(5.8,-1.0) node[inner sep=0pt, opacity=0.9] (img1) { \includegraphics[width=2.8cm,height=2.8cm]{FIG-crop-2270.png} };
\begin{scope}[xshift=2.86cm,yshift=-4.1cm, scale=0.7]
   %-----------Box1
    \draw[white=0.9,thick] (2.25,5.65) rectangle (3.2,5.15);
  \draw[white,fill=white] (2.25,5.65) circle (.3ex);
 \draw[white,fill=white] (3.2,5.15) circle (.3ex);
 \path[->](+2.8,5.85) node[white,very thick, inner sep=1pt]  {\scriptsize $r_1$};
  
    %-----------Box2
  \draw[white=0.9,thick] (+4.6,4.5) rectangle (+5.8,2.82);
   \draw[white,fill=white] (+4.6,4.5) circle (.3ex);
 \draw[white,fill=white] (+5.8,2.82) circle (.3ex);
  \path[->](+5.2,4.7) node[white,very thick, inner sep=1pt]  {\scriptsize $r_2$};
\end{scope}

\path[->](0.0,-3.55) node[black,very thick,inner sep=0pt] {\small (a)};
\path[->](2.9,-3.55) node[black,very thick,inner sep=0pt] {\small (b)};
\path[->](5.8,-3.55) node[black,very thick,inner sep=0pt] {\small (c)};

\end{tikzpicture}
\caption{Example of (a) two overlapped regions from the same category that must not be merged, and of (b) three detected regions in which (c) two of them were merged using their confidence score to define the new bounding box dimensions.} 
\label{fig:nms}
\end{figure}

Each detector in the ensemble runs separately on each block of its input image (see Figure~\ref{fig:framework}(f)), and we end up with multiple results per block that must be merged into a single outcome. Before that, we eliminate regions whose confidence value is below a threshold $t$. Table~\ref{tab:parameters_table2} indicates the value of $t$ for each detector. Non-discarded regions for all blocks are mapped back to the ROI coordinate space. Many vehicles may be detected multiple times, either by being detected by different models of the ensemble or by appearing in overlapped block regions. The Non-Maximum Suppression (NMS)~\cite{Felzenszwalb2010} algorithm with minor adaptations is used to discard duplicate regions belong to the same object. Consider $\bar{\mathcal{R}} = \{\bar{r}_1, \bar{r}_2, \dots\}$ the set of regions not yet filtered by NMS. NMS selects the region $\bar{r}_i \in \bar{\mathcal{R}}$ with the highest confidence score and loops through $\bar{\mathcal{R}}$ looking for other regions $\bar{r}_j$ that have an Intersection over Union (IoU) greater than a given threshold $\sigma$:
\begin{equation}
 \textrm{IoU}(b(\bar{r}_i), b(\bar{r}_j)) = \frac{\textrm{area}(b(\bar{r}_i) \cap b(\bar{r}_j))}{\textrm{area}(b(\bar{r}_i) \cup b(\bar{r}_j))} > \sigma
 \label{eq1}
\end{equation}
The IoU metric takes into account the total area of both analyzed regions, which is particularly interesting for satellite imagery, where in many cases a significant intersection of objects does not mean that they should be merged (see Figure~\ref{fig:nms}(a)). The region $\bar{r}_i $ and all other regions that satisfy Equation~\ref{eq1} will form a subset $\bar{\mathcal{R}}_k \subseteq \bar{\mathcal{R}}$ that will be merged into a single region $r_i$:
\begin{equation}
  {r_i}(b) = \frac{\displaystyle \sum_{\bar{r} \in \bar{\mathcal{R}}_k} w(\bar{r}) \times b(\bar{r})}{\displaystyle \sum_{\bar{r} \in \bar{\mathcal{R}}_k} w(\bar{r})} 
  \end{equation}
with $r_i$ being part of the final result $\mathcal{R}$. As may be seen, instead of discarding overlapped regions with lower confidence score, we combine all regions in $\bar{\mathcal{R}}_k$ through a weighted average. This avoids noise in the final bounding box coordinates, as shown in Figures~\ref{fig:nms}(b)~and~\ref{fig:nms}(c). $\bar{\mathcal{R}}_k$ is then removed from $\bar{\mathcal{R}}$ and the process is repeated until $\bar{\mathcal{R}}$ is empty.

\begin{figure*}[!htb]
  \centering
  \includegraphics[width=1.0\linewidth]{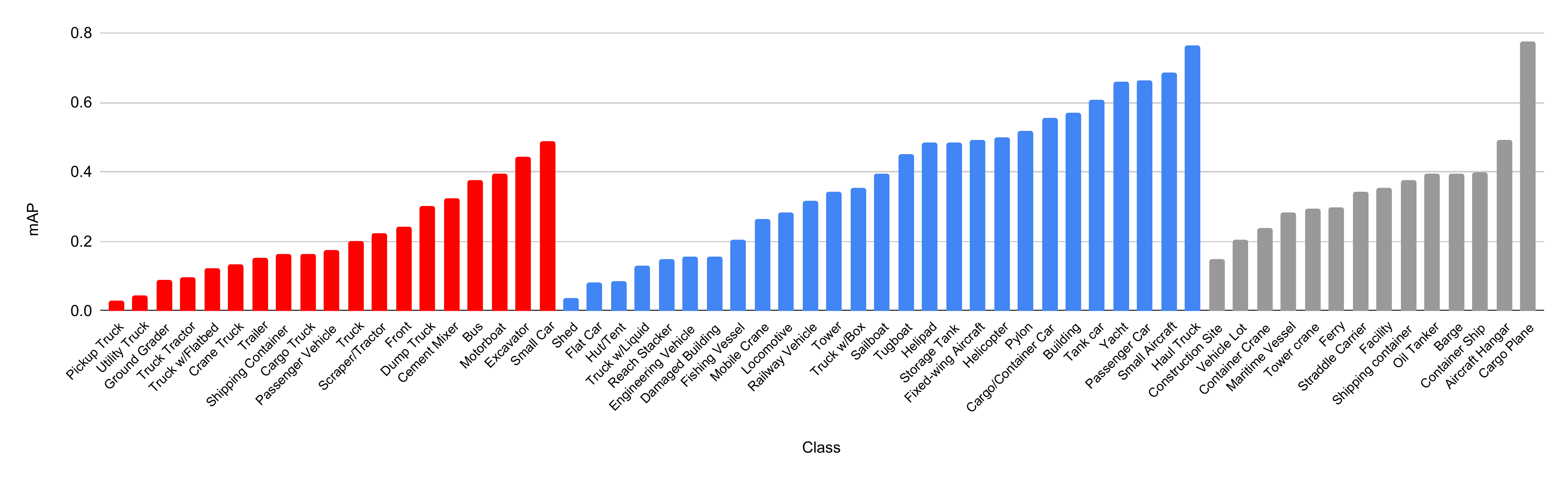}
  \vspace{-0.8cm}
  \caption{Mean average precision per xView class. Red, blue and gray bars represent small, medium and large targets, respectively.}
  \label{fig:detection} 
\end{figure*}

An example of the final output for a ROI is illustrated in Figure~\ref{fig:framework}(g). This module of our framework can be updated in the future to incorporate recent state-of-the-art solutions that address some recurring challenges in satellite data, such as class unbalancing~\cite{lin2020}, scale variations~\cite{Zhou2018}, and unrealistic false alarms~\cite{Hu2018}, as a way to increase the detection accuracy or reduce the ensemble size.

\subsection{Temporal analysis}
Given enough time and acquisition frequency, it is possible to apply time series analysis techniques~\cite{Shumway2005} to learn the standard behavior-patterns in each ROI and then identify trends in these areas. It is vital to cope with typical variations in different scales, such as seasonal variations along the year, monthly variations, weekly variations, or even daily variations. Recent approaches based on concept drift~\cite{Gama2014} can help to identify a change in behavior while avoiding outlier data. Depending on the amount of data available for training, one can also explore the use of recurrent neural networks for time-series forecasting~\cite{Lai2018,Qin2017} followed by the detection of abnormal behaviors. However, this module has not been implemented yet. We need access to appropriate data to study these variations. 

%%%%%%%%%%%%%%%%%%%%%%%%%%%%%%%%%%%%%%%%%%%%%%%%%%%%%%%%%%%%%%%%%%%%%%%%%%%%%%%%
\section{Experiments}\label{sec:results}
\subsection{Detector evaluation}\label{sec:exp_detector}
To evaluate the detection stage, we used the xView dataset~\cite{xview2018}. It contains a training set with 847 high-resolution images ($0.3$ GSD) and about one million annotations for 60 classes of objects. It also contains an evaluation subset with 282 images to which no annotations were provided, and a sequestered testing subset with 284 images. Image sizes range from $2564 \times 2576$ to $3187 \times 4994$ pixels. The interpolated mean average precision (mAP), detailed by Henderson and Ferrari~\cite{DBLP:journals/corr/HendersonF16}, can be computed for the training set through its object annotations. A mAP value for the entire evaluation set could be obtained in an online submission system while the xView competition was running. The precision for the sequestered testing set could only be computed by the xView organizers.

The xView classes were divided into three groups according to the object size: small, medium and large. A complete list of classes per group is available in Figure~\ref{fig:detection}. The ensemble configuration described in Section~\ref{sec:car-detection} is used for the small group only, which includes the class \emph{`Small Car'} that we use for vehicle detection. A complete description of our ensemble is shown in Table~\ref{tab:parameters_table2}, including additional detectors and their parameters for objects of medium and large sizes as well. As other classes are useful for future analyses (discussed in Section~\ref{sec:discussion}), we report detection results for them as well.

\begin{table}[!htb]
\renewcommand*{\arraystretch}{1.2}
\caption{Parameters of our ensemble of SSDs for all classes in the xView dataset. Detectors \#1-\#3 are used for small objects; detectors \#1-\#4 for medium objects; and detectors \#3-\#5 for large objects.}
\label{tab:parameters_table2}
\begin{center}
\begin{tabular}{|c||c|c|c|c|c|}
\hline
& Scale & Overlap & Thr. & Model & Size group \\ \hline \hline
Det. \#1 & 1.0 & 0px & 0.15 & Vanilla & Small\&Medium \\ \hline
Det. \#2 & 1.3 & 0px & 0.06 & Vanilla & Small\&Medium \\ \hline
Det. \#3 & 1.0 & 100px & 0.06 & Multires & All \\ \hline
Det. \#4 & 0.7 & 100px & 0.5 & Multires & Medium\&Large \\ \hline
Det. \#5 & 0.6 & 0px & 0.06 & Multires & Large \\ \hline
\end{tabular}
\end{center}
\end{table}

In Figure~\ref{fig:detection} we show the mAP per class of our ensemble for the training set, which is the only set with annotations that allowed us to do so. Even though the baseline detectors used in our ensemble were trained with this set, the problem is hard enough to prevent detectors from reaching perfect accuracy. Still, this figure gives a good idea of which classes are more accurate than others. The class \emph{`Small Car'}, for instance, reaches nearly 0.5 mAP.

This ensemble was submitted to the xView competition and achieved a mAP of 29.88 in the evaluation set, while the baselines \emph{Vanilla} and \emph{Multires} achieved 20.87 and 18.14, respectively. It ranked third over all contestants in the sequestered testing set (see Table~\ref{tab:xview_leaderboard}), evidencing the potential of the approach for detecting targets in satellite images. Our framework can process up to ten image blocks per second on a modern GPU, which allows updating the object count of thousands of ROIs per hour. This pace is more than enough to handle city-scale applications, even if image acquisition occurs on a daily basis.

\begin{table}[!htb]
\renewcommand*{\arraystretch}{1.2}
\caption{Final xView leaderboard: mAP per size group and overall.}
\label{tab:xview_leaderboard}
\begin{center}
\begin{tabular}{|c|c|c|c||c|}
    \hline
Rank & \multicolumn{4}{c|}{mAP} \\ \cline{2-5}
Country & Small & Medium & Large & Score \\ \hline
1 (Russia) & 0.1965 & 0.3371 & 0.3400 & 0.2932 \\ \hline
2 (Australia) & 0.1632 & 0.3595 & 0.2536 & 0.2727 \\ \hline
3 (\textbf{proposed/USA}) & 0.1733 & 0.3261 & 0.3039 & 0.2726 \\ \hline
4 (Italy) & 0.1680 & 0.3339 & 0.2821 & 0.2693 \\ \hline
5 (USA) & 0.1587 & 0.2657 & 0.2511 & 0.2284 \\ 
\hline
\end{tabular}
\end{center}
\end{table}

\subsection{Temporal evaluation}\label{sec:exp_temporal}

The fMoW dataset~\cite{christie2018functional} contains more than one million excerpts of satellite images split into training, evaluation, and testing subsets. Even though it provides high-resolution pan-sharpened images~\cite{li2015}, most of them do not have a GSD as low as the ones in the xView dataset. This because this dataset was created for land use classification, not for small object detection. However, it provides temporal views of the same region, which is very interesting for this experiment. Temporality brings variations in shadows, viewpoints, weather, and vehicles in the scene, the last being our primary focus.

\begin{figure*}[!htb]

\subfigure[The difference between the lowest and the highest vehicle count is 12 in the ground truth and 25 in our count]{\label{fig:temporal1}
 \begin{tikzpicture}
  \renewcommand*{\arraystretch}{0.7}
  \draw(0.0,0.0) node[inner sep=0pt] (img1) {
   \includegraphics[height=7.0cm,width=0.98\linewidth]{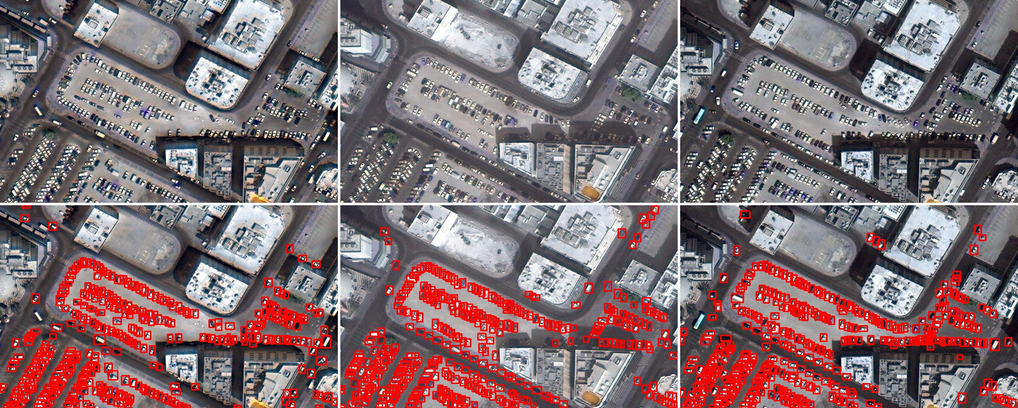}};
   
   \path[->](-4.25,+0.38) node[black,very thick,fill=white,fill opacity=0.98, inner sep=0pt]  {\begin{tabular}{l}\textbf{\scriptsize Ground truth:}\\\scriptsize small-car = 336 \end{tabular}};

   \path[->](-4.3,-3.15) node[black,very thick,fill=white,fill opacity=0.98, inner sep=0pt]  {\begin{tabular}{l}\textbf{\scriptsize Detection result:}\\\scriptsize small-car = 410 \end{tabular}};
   
   \path[->](1.70,+0.38) node[black,very thick,fill=white,fill opacity=0.98, inner sep=0pt]  {\begin{tabular}{l}\textbf{\scriptsize Ground truth:}\\\scriptsize small-car = 344 \end{tabular}};

   \path[->](1.65,-3.15) node[black,very thick,fill=white,fill opacity=0.98, inner sep=0pt]  {\begin{tabular}{l}\textbf{\scriptsize Detection result:}\\\scriptsize small-car = 418 \end{tabular}};
   
   \path[->](7.65,+0.38) node[black,very thick,fill=white,fill opacity=0.98, inner sep=0pt]  {\begin{tabular}{l}\textbf{\scriptsize Ground truth:}\\\scriptsize small-car = 332 \end{tabular}};

   \path[->](7.60,-3.15) node[black,very thick,fill=white,fill opacity=0.98, inner sep=0pt]  {\begin{tabular}{l}\textbf{\scriptsize Detection result:}\\\scriptsize small-car = 435 \end{tabular}};

 \end{tikzpicture}
}

\subfigure[The difference between the lowest and the highest vehicle count is 81 in the ground truth and 103 in our count]{\label{fig:temporal2}
 \begin{tikzpicture}
  \renewcommand*{\arraystretch}{0.7}
  \draw(0.0,0.0) node[inner sep=0pt] (img1) {
   \includegraphics[height=7.0cm,width=0.98\linewidth]{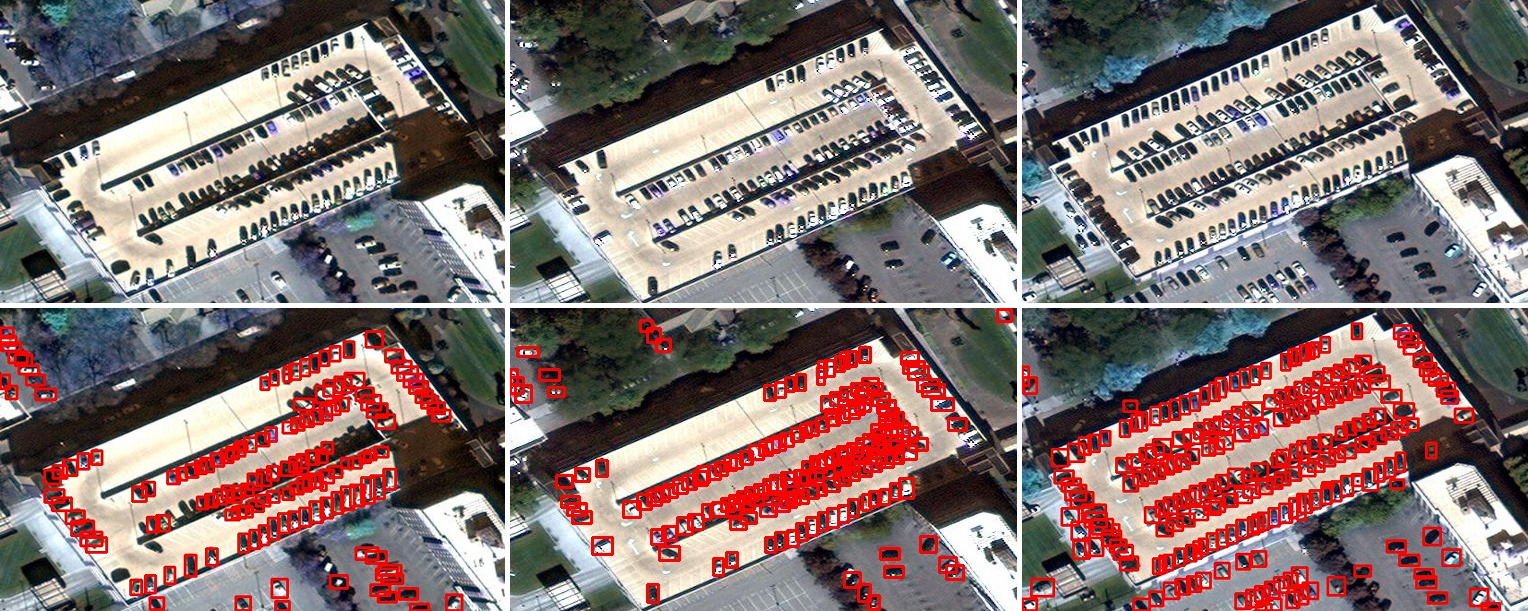}};

   \path[->](-4.25,+0.38) node[black,very thick,fill=white,fill opacity=0.98, inner sep=0pt]  {\begin{tabular}{l}\textbf{\scriptsize Ground truth:}\\\scriptsize small-car = 154 \end{tabular}};

   \path[->](-4.3,-3.15) node[black,very thick,fill=white,fill opacity=0.98, inner sep=0pt]  {\begin{tabular}{l}\textbf{\scriptsize Detection result:}\\\scriptsize small-car = 150 \end{tabular}};
   
   \path[->](1.70,+0.38) node[black,very thick,fill=white,fill opacity=0.98, inner sep=0pt]  {\begin{tabular}{l}\textbf{\scriptsize Ground truth:}\\\scriptsize small-car = 163\end{tabular}};

   \path[->](1.65,-3.15) node[black,very thick,fill=white,fill opacity=0.98, inner sep=0pt]  {\begin{tabular}{l}\textbf{\scriptsize Detection result:}\\\scriptsize small-car = 194 \end{tabular}};
   
   \path[->](7.65,+0.38) node[black,very thick,fill=white,fill opacity=0.98, inner sep=0pt]  {\begin{tabular}{l}\textbf{\scriptsize Ground truth:}\\\scriptsize small-car = 235 \end{tabular}};

   \path[->](7.60,-3.15) node[black,very thick,fill=white,fill opacity=0.98, inner sep=0pt]  {\begin{tabular}{l}\textbf{\scriptsize Detection result:}\\\scriptsize small-car = 253 \end{tabular}};

 \end{tikzpicture}
}

\subfigure[The difference between the lowest and the highest vehicle count is 702 in the ground truth and 720 in our count]{\label{fig:temporal3}
 \begin{tikzpicture}
  \renewcommand*{\arraystretch}{0.7}
  \draw(0.0,0.0) node[inner sep=0pt] (img1) {
   \includegraphics[height=7.0cm,width=0.98\linewidth]{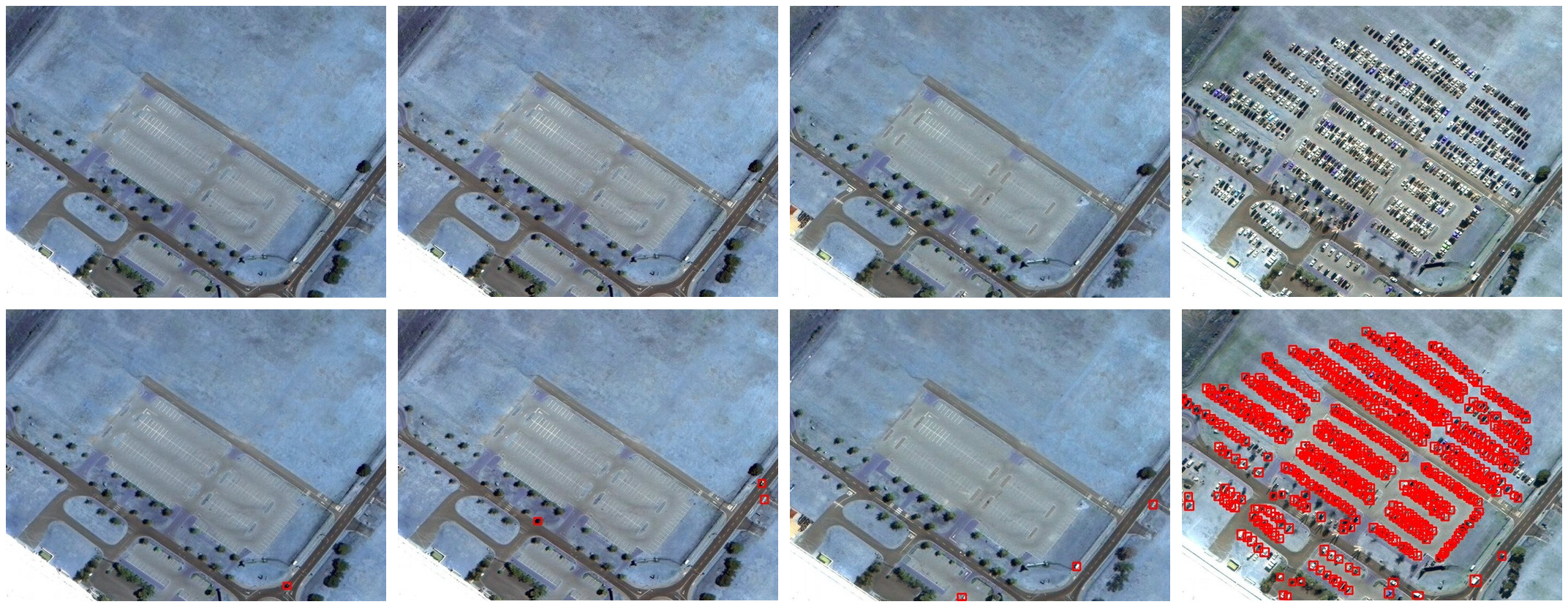}};

   \path[->](-5.75,+0.40) node[black,very thick,fill=white,fill opacity=0.98, inner sep=0pt]  {\begin{tabular}{l}\textbf{\scriptsize Ground truth:}\\\scriptsize small-car = 1 \end{tabular}};

   \path[->](-5.8,-3.10) node[black,very thick,fill=white,fill opacity=0.98, inner sep=0pt]  {\begin{tabular}{l}\textbf{\scriptsize Detection result:}\\\scriptsize small-car = 1 \end{tabular}};
   
   \path[->](-1.28,+0.40) node[black,very thick,fill=white,fill opacity=0.98, inner sep=0pt]  {\begin{tabular}{l}\textbf{\scriptsize Ground truth:}\\\scriptsize small-car = 1 \end{tabular}};
   
   \path[->](-1.33,-3.10) node[black,very thick,fill=white,fill opacity=0.98, inner sep=0pt]  {\begin{tabular}{l}\textbf{\scriptsize Detection result:}\\\scriptsize small-car = 3 \end{tabular}};
   
   \path[->](3.19,+0.40) node[black,very thick,fill=white,fill opacity=0.98, inner sep=0pt]  {\begin{tabular}{l}\textbf{\scriptsize Ground truth:}\\\scriptsize small-car = 1 \end{tabular}};
   
   \path[->](3.12,-3.10) node[black,very thick,fill=white,fill opacity=0.98, inner sep=0pt]  {\begin{tabular}{l}\textbf{\scriptsize Detection result:}\\\scriptsize small-car = 3 \end{tabular}};
   
   \path[->](7.65,+0.40) node[black,very thick,fill=white,fill opacity=0.98, inner sep=0pt]  {\begin{tabular}{l}\textbf{\scriptsize Ground truth:}\\\scriptsize small-car = 703 \end{tabular}};
    
   \path[->](7.6,-3.10) node[black,very thick,fill=white,fill opacity=0.98, inner sep=0pt]  {\begin{tabular}{l}\textbf{\scriptsize Detection result:}\\\scriptsize small-car = 721\end{tabular}};

 \end{tikzpicture}
}

\caption{Detection results for multiple samples of a same region with a (a) small, (b) medium, and (c) large variation in the flow of vehicles. The samples are sorted from left to right in a non-decreasing order of the number of detected vehicles.}
\label{fig:temporal} 
\end{figure*}
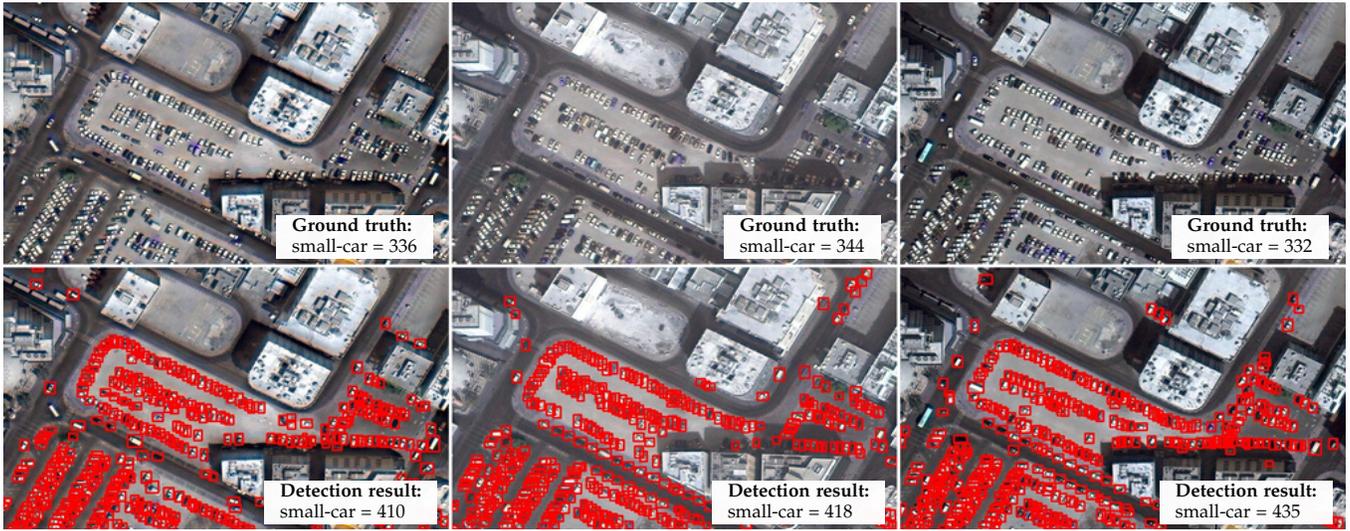
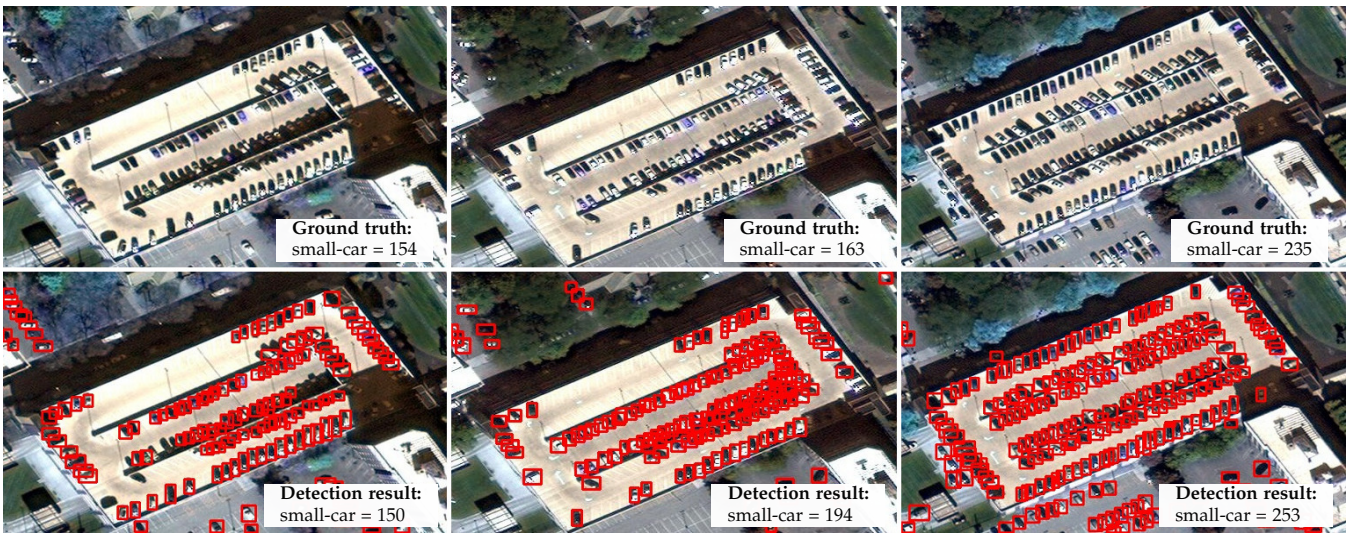
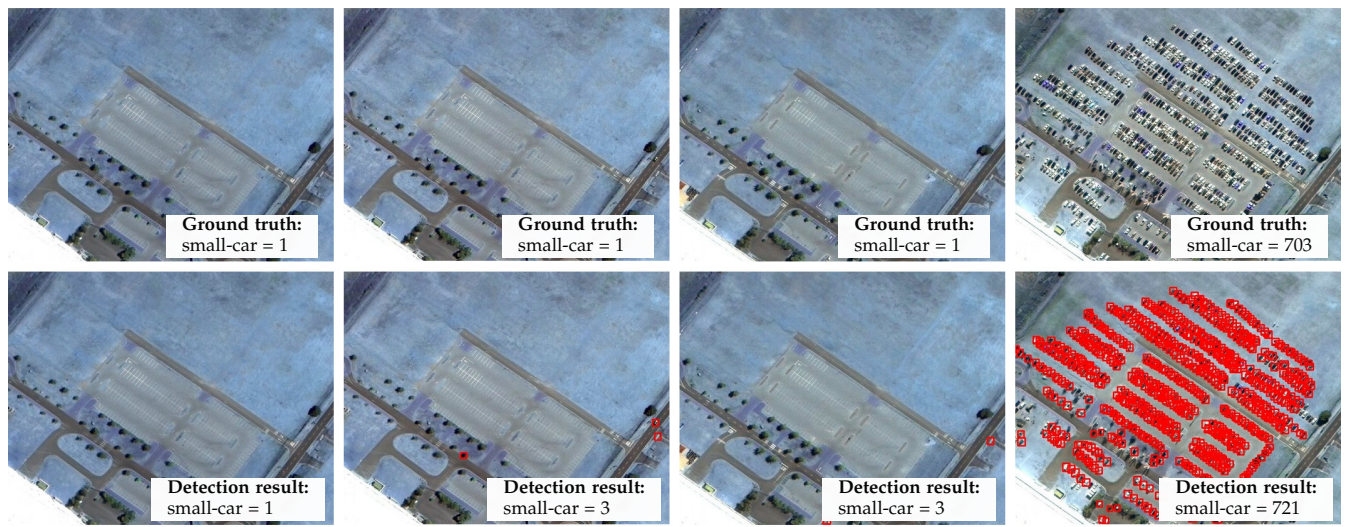

Each region in the dataset represents one of the 63 categories, including a false detection category that aggregates different types of regions that do not fit into the other 62 defined categories. Among those classes, we are particularly interested in one: \emph{`parking\_lot\_or\_garage'}. We looked for regions of this class that:
\begin{itemize}
    \item have three or more samples with GSD smaller than $0.4$ and dimensions greater than $300\times300$ pixels (one block in Section~\ref{sec:car-detection}); and
    \item show an open-air parking lot (as \emph{`parking\_lot\_or\_garage'} includes closed garage buildings).
\end{itemize}
Following these criteria, we ended up with nine regions with three to nine images each. To improve the detection outcome in the regions with lower resolution, we manually upsampled them so that their objects' sizes looked closer to how they were supposed to look on xView images (i.e., GSD $\approx 0.3$). We also increased the confidence threshold to $0.25$ to eliminate false positives caused by the upsampling operation ({\it e.g.}, edge blurring and noise amplification). We ran our vehicle detector using each of these regions as our ROI, and even though there are false positives and negatives in nearly all regions, the count is consistent enough for further automated analyses. We sorted samples of the same region in a non-decreasing order of the number of detected vehicles to illustrate the potential of the framework to perceive gradual changes in the flow of vehicles.
Three regions with small, medium and high variation in the number of vehicles are respectively shown in Figures~\ref{fig:temporal1},~\ref{fig:temporal2}~and~\ref{fig:temporal3}. One can argue if the sorting is correct or not for some of the samples, but the overall quality of the process is evident.

Figure~\ref{fig:temporal1} shows the region with the lowest variation in the flow of vehicles. As can be observed, we can identify regions with a stable count even when there is a high volume of cars. This ability is essential for stay-at-home enforcement when a decrease is expected but is not confirmed, requiring further action from the authorities ({\it e.g.}, suspension of activities in public, social, and private sectors). Figures~\ref{fig:temporal2}~and~\ref{fig:temporal3} respectively show regions with medium and high variation. Recognizing such changes is important in both directions, either increasing or decreasing. For instance, a decreasing vehicle count in hospitals can point out a reduction in the outbreak, and in airports can indicate a reduction in the economic activity. Meanwhile, a similar trend in residential areas can suggest both an outbreak reduction or stay-at-home disobedience, depending on the context. Finally, an increasing vehicle count can reveal critical regions that require more attention from authorities. Sudden increases in supermarkets can detect panic buying, and in convention centers the occurrence of large unauthorized events (see Figure~\ref{fig:temporal3}).

We manually annotated more than 2,000 cars in the images shown in Figure~\ref{fig:temporal} using an open-source tool\footnote{https://github.com/tzutalin/labelImg} to compare our detection results with the ground truth. Each annotation is an axis-aligned rectangle delimited by its top-left and bottom-right corners and categorized as one of the 60 xView object classes. Our detector achieved a $0.59$ mAP for small cars in these images, which is on par with the accuracy on xView. The Mean Absolute Percentage Error (MAPE) in car counting for images with more than 100 annotations is approximately 15\%, which indicates that the number of detections and annotations are relatively close to each other. Although these count values have a larger deviation in some regions (see Figure~\ref{fig:temporal1}), the error tends to be similar in images of the same region. As a result, we can estimate the amount of change in the number of vehicles accurately. With this information, we can devise indicators for decision-makers using different ROI groups and their expected behavior ({\it e.g.}, a stable count in supermarkets, a decrease in schools, an increase in rental car facilities). Besides that, with proper temporal sequences of images from local businesses' parking lots, we could quantify the impact of COVID-19 on their earnings~\cite{zsolt2018}. Finally, when ROIs have geographic coordinates, it is possible to interpolate these estimates to neighboring areas and produce heatmaps, as illustrated in Figure~\ref{fig:framework}(h).

%%%%%%%%%%%%%%%%%%%%%%%%%%%%%%%%%%%%%%%%%%%%%%%%%%%%%%%%%%%%%%%%%%%%%%%%%%%%%%%%
\subsection{COVID-19 case studies}
\label{sec:casestudies}

\begin{figure*}[!htb]
   \renewcommand*{\arraystretch}{0.7}
   \subfigure[North Korean commercial vessels are idled in their home ports after the COVID-19 outbreak]{
    \begin{tikzpicture}
    %\path[->](0.0,2.5) node[black, inner sep=0pt]  {Input image};
    %\path[->](8.9,2.5) node[black, inner sep=0pt]  {Detection/classification (proposed framework)};
        \draw(0.0,0.0) node[inner sep=0pt] (img1) {
     \includegraphics[height=4.6cm,width=8.75cm]{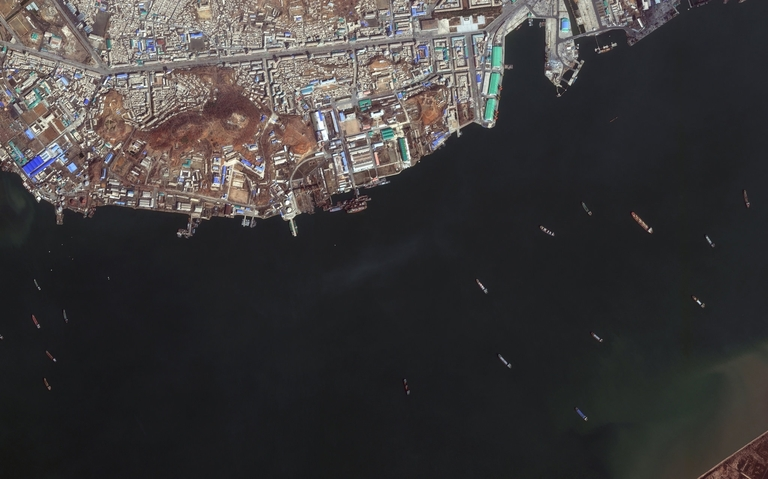}}; 
     \path[->](3.10,-1.90) node[black,very thick,fill=white,fill opacity=0.98, inner sep=0pt]  {\begin{tabular}{l}\textbf{\scriptsize Ground truth:}\\\scriptsize vessels (sea) = 24 \end{tabular}};
     
        \draw(8.9,0.0) node[inner sep=0pt] (img1) {
     \includegraphics[height=4.6cm,width=8.75cm]{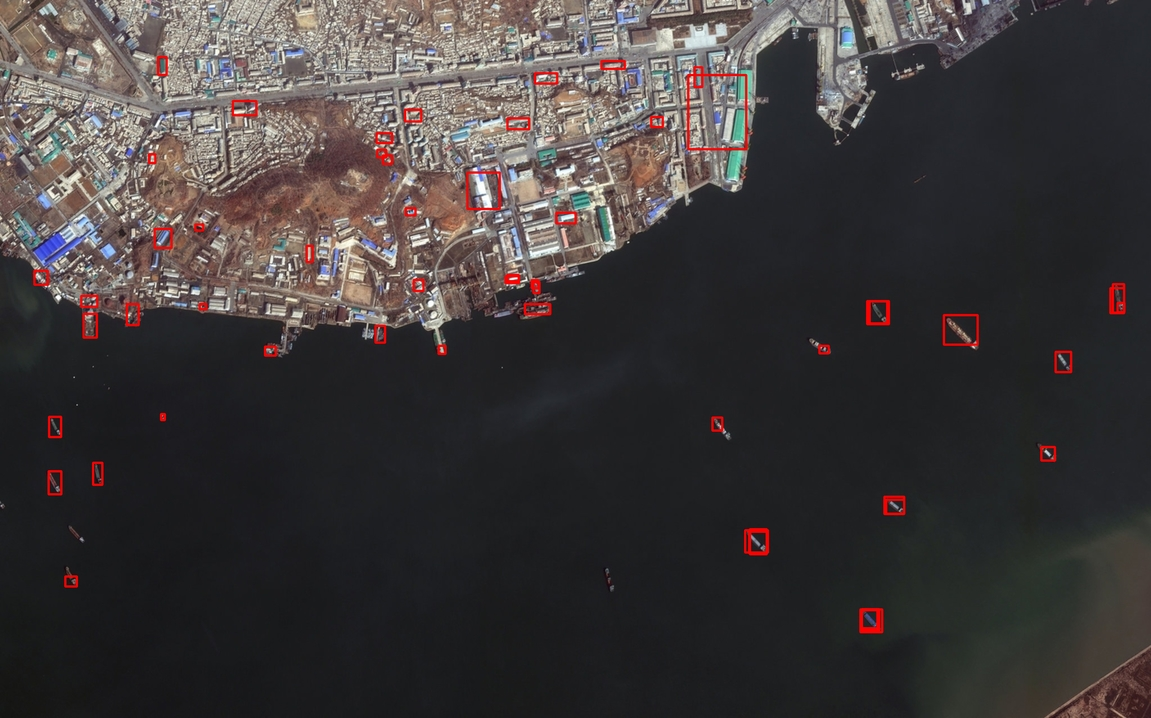}}; 
       \path[->](11.88,-1.20) node[black,very thick,fill=white,fill opacity=0.98, inner sep=0pt]  {\begin{tabular}{l}\textbf{\scriptsize Detection result:}\\\scriptsize motorboat = 12\\\scriptsize fishing vessel = 6 \\\scriptsize sailboat = 1 \\\scriptsize tugboat = 2 \\\scriptsize maritime vessel = 1\\\scriptsize \textbf{...}\end{tabular}};
     \end{tikzpicture}
   }\\ 
    \subfigure[The construction of a hospital outside Moscow (Russia) to treat COVID-19 cases]{
    \begin{tikzpicture}
        \draw(0.0,0.0) node[inner sep=0pt] (img1) {
     \includegraphics[height=4.6cm,width=8.75cm]{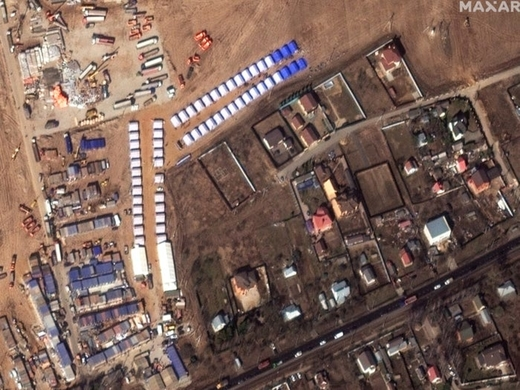}}; 
        \draw(2.85,-1.3) node[inner sep=0pt] (img1) {
     \includegraphics[height=2.0cm,width=3.0cm]{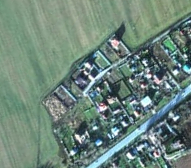}}; 
     
     \path[->](2.58,+1.80) node[black,very thick,fill=white,fill opacity=0.98, inner sep=0pt]  {\begin{tabular}{l}\textbf{\scriptsize Ground truth:}\\\scriptsize truck (several models) = 28 \end{tabular}};
     
     \path[->](2.85,-0.2) node[black,very thick,fill=white,fill opacity=0.98, inner sep=1pt]  {\footnotesize Before (low resolution)};
        \draw(8.9,0.0) node[inner sep=0pt] (img1) {
     \includegraphics[height=4.6cm,width=8.75cm]{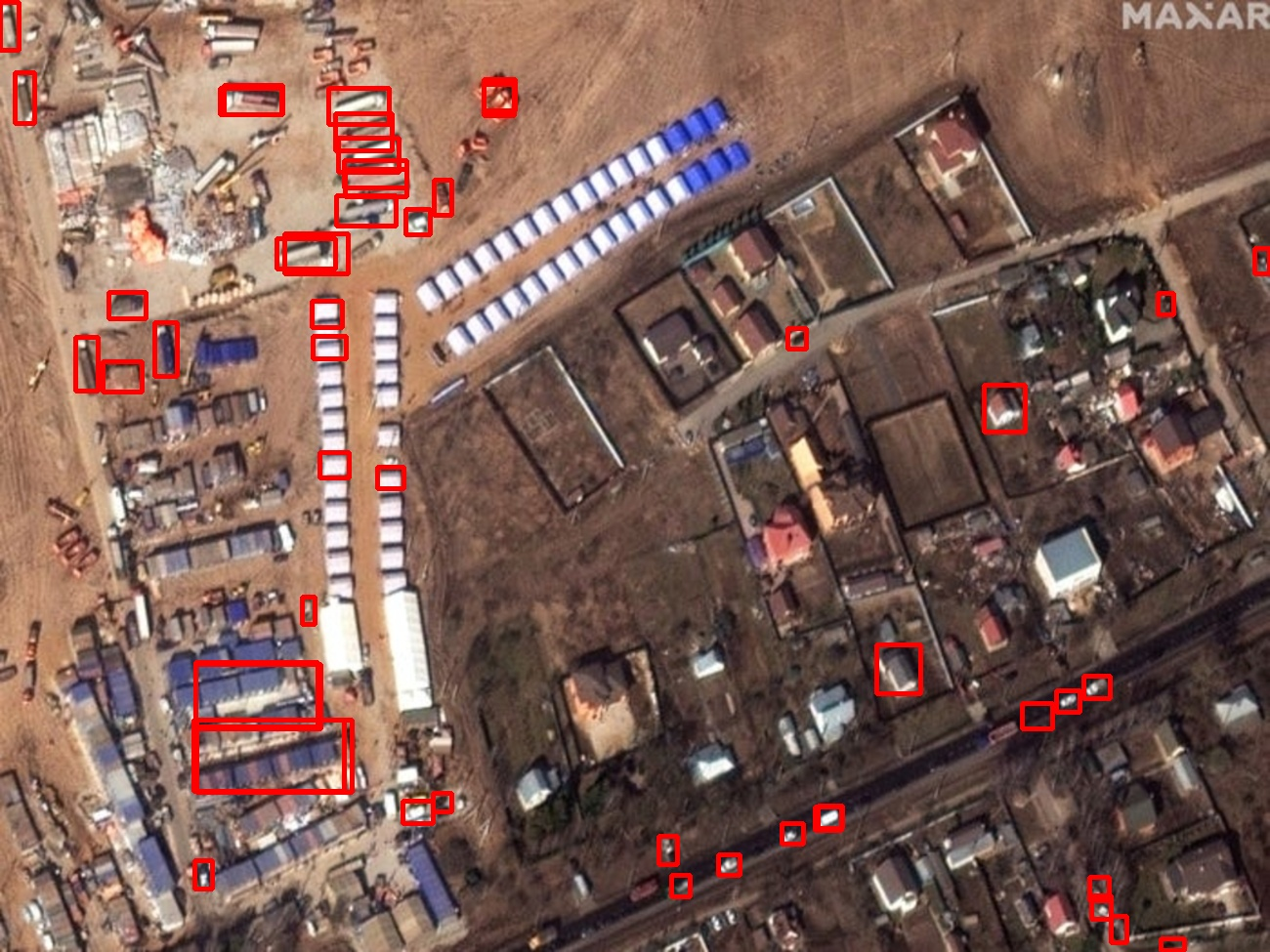}}; 
        \path[->](11.94,-1.50) node[black,very thick,fill=white,fill opacity=0.98, inner sep=0pt]  {\begin{tabular}{l}\textbf{\scriptsize Detection result:}\\\scriptsize truck = 17\\\scriptsize truck w/Box = 2 \\\scriptsize mobile crane = 6\\\scriptsize \textbf{...}\end{tabular}};
     \end{tikzpicture}
   }
   \\ 
    \subfigure[COVID-19 drive-through testing facility built at Munich in Germany]{
    \begin{tikzpicture}
     \draw(0.0,0.0) node[inner sep=0pt] (img1) {
     \includegraphics[height=4.6cm,width=8.75cm]{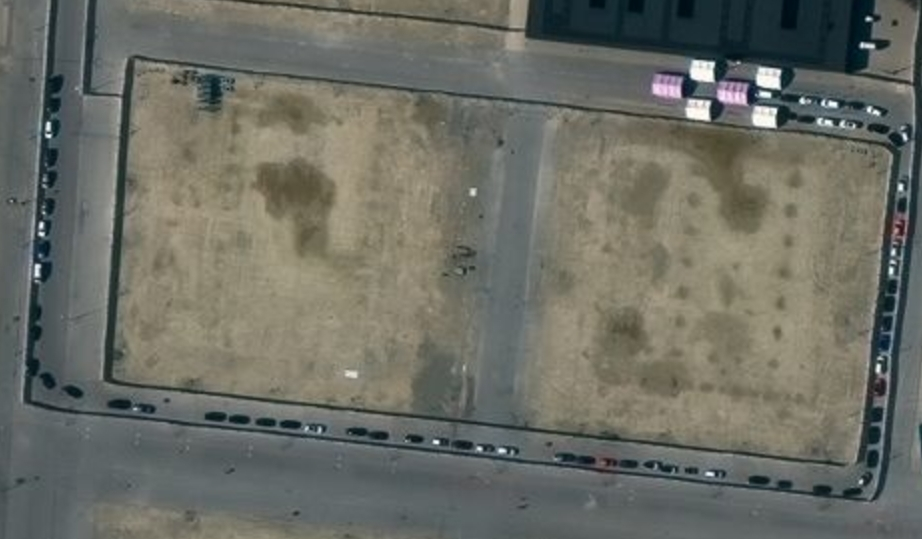}}; 
     \path[->](3.10,-1.90) node[black,very thick,fill=white,fill opacity=0.98, inner sep=0pt]  {\begin{tabular}{l}\textbf{\scriptsize Ground truth:}\\\scriptsize small-car = 68 \end{tabular}};
     
     \draw(8.9,0.0) node[inner sep=0pt] (img1) {
     \includegraphics[height=4.6cm,width=8.75cm]{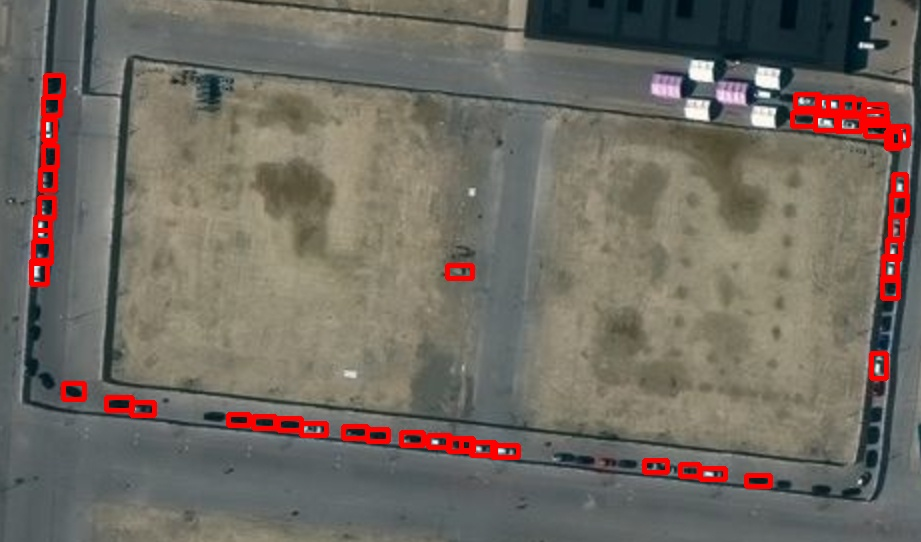}}; 
        \path[->](11.96,-1.77) node[black,very thick,fill=white,fill opacity=0.98, inner sep=0pt]  {\begin{tabular}{l}\textbf{\scriptsize Detection result:}\\\scriptsize small-car = 49 \\\scriptsize \textbf{...}\end{tabular}};
     \end{tikzpicture}
   }
   \caption{Satellite imagery from world scenes related to the COVID-19 pandemic and statistics about vehicles/infrastructure available. \textbf{Photo credit}: \textsc{Satellite image 2020 Maxar Technologies}.}
   \label{fig:covid-scenarios-a} 
\end{figure*}

The results presented so far show the potential of the proposed framework to address the intended problem, but miss the real thing -- the COVID-19  outbreak.
To evaluate our framework in real-world situations, we collected satellite imagery released to the press to illustrate the impact of the COVID-19 over the globe. It is worth noting that these are not always raw, high-resolution images, such as those provided by xView. Many of these images had artifacts that impair automated analysis, like watermarks and low-resolution.
We followed the same procedure employed for fMoW images, and manually upsampled images so that the GSD is approximately $0.3$ and increased the confidence threshold to $0.25$.

\begin{figure*}[!htb]
   \centering
   \renewcommand*{\arraystretch}{0.7}
   \subfigure[Phoenix airport (USA) parking-lot before COVID-19 outbreak]{
     \begin{tikzpicture}
        \draw(0.0,0.0) node[inner sep=0pt] (img1) {
     \includegraphics[height=8.6cm,width=8.97cm]{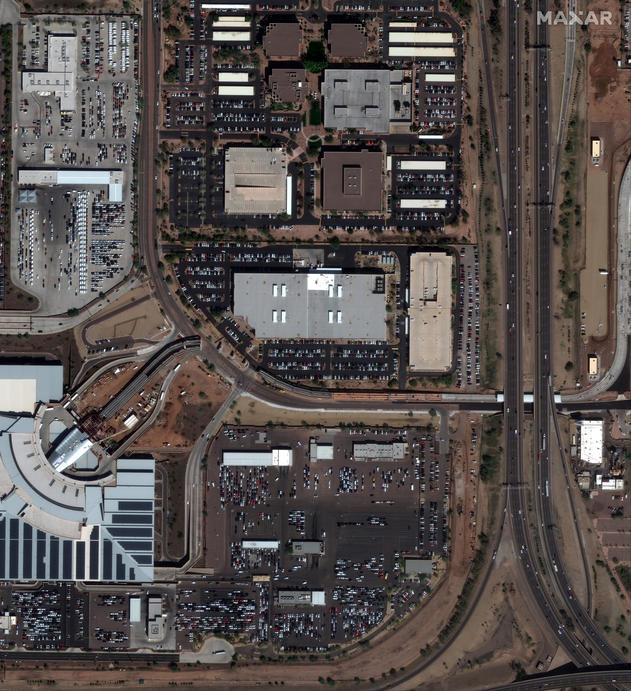}}; 
        \draw(9.15,0.0) node[inner sep=0pt] (img1) {
     \includegraphics[height=8.6cm,width=8.97cm]{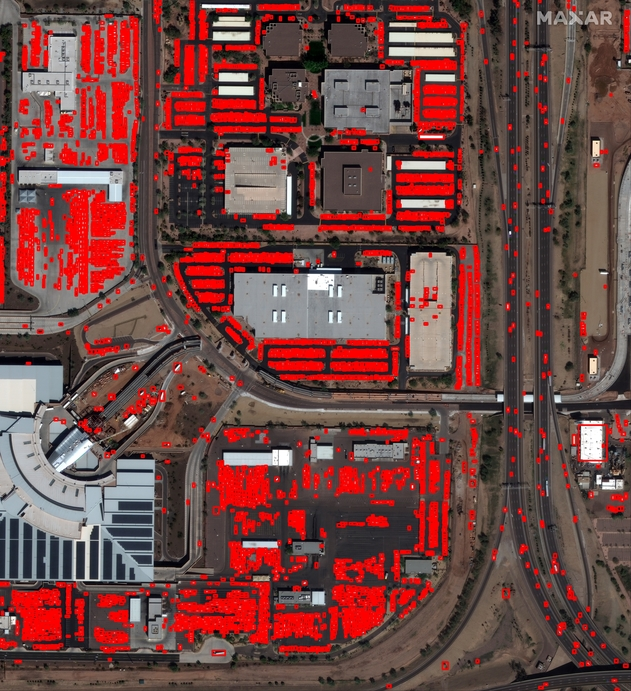}}; 
       \path[->](3.16,-3.75) node[black,very thick,fill=white,fill opacity=0.98, inner sep=0pt]  {\begin{tabular}{l}\textbf{\scriptsize Ground truth:}\\\scriptsize small-car = 5336  \\\scriptsize \textbf{...}\end{tabular}};
       \path[->](12.25,-3.75) node[black,very thick,fill=white,fill opacity=0.98, inner sep=0pt]  {\begin{tabular}{l}\textbf{\scriptsize Detection result:}\\\scriptsize small-car = 5989 \\\scriptsize \textbf{...}\end{tabular}};
     \end{tikzpicture}
   } \\
   \subfigure[Phoenix airport (USA) parking-lot after COVID-19 outbreak]{
     \begin{tikzpicture}
        \draw(0.0,0.0) node[inner sep=0pt] (img1) {
     \includegraphics[height=8.6cm,width=8.97cm]{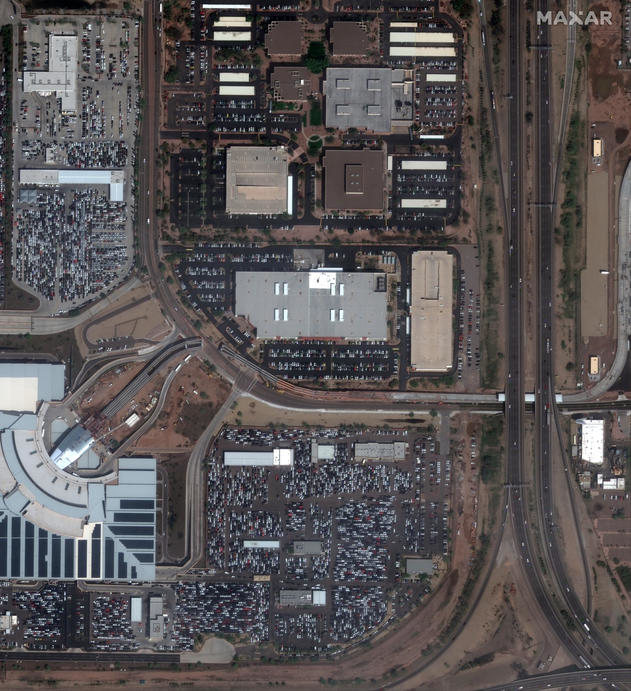}}; 
        \draw(9.15,0.0) node[inner sep=0pt] (img1) {
     \includegraphics[height=8.6cm,width=8.97cm]{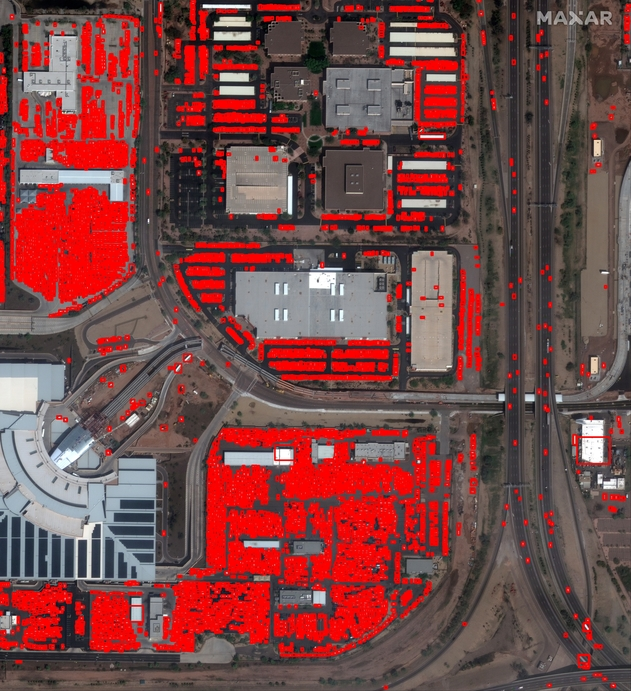}}; 
       \path[->](3.16,-3.75) node[black,very thick,fill=white,fill opacity=0.98, inner sep=0pt]  {\begin{tabular}{l}\textbf{\scriptsize Ground truth:}\\\scriptsize small-car = 8299 \\\scriptsize \textbf{...}\end{tabular}};
       \path[->](12.25,-3.75) node[black,very thick,fill=white,fill opacity=0.98, inner sep=0pt]  {\begin{tabular}{l}\textbf{\scriptsize Detection result:}\\\scriptsize small-car = 7717  \\\scriptsize \textbf{...}\end{tabular}};
     \end{tikzpicture}
   }
   \caption{Satellite imagery from world scenes before and after the COVID-19 outbreak and statistics about vehicles/infrastructure available. \textbf{Photo credit}: \textsc{Satellite image 2020 Maxar Technologies}.}
   \label{fig:covid-scenarios-b1} 
\end{figure*}

Repurposing the framework for other economic activities is simple. The places to monitor, i.e. roads, ports, etc., can be changed easily. The objects of interest can be changed, and our detector can handle many different options. See Figure~\ref{fig:detection} for a complete list. For instance, Figure~\ref{fig:covid-scenarios-a}(a) shows an image from North Korean commercial vessels used to transport coal and other commodities. According to Christoph Koettl from the New York Times\footnote{https://www.nytimes.com/2020/03/26/video/coronavirus-north-korea.html}, they stopped in their home ports as an attempt to prevent the virus transmission. We did not find an image with a reasonable resolution before the outbreak. Still, according to other satellite data, this concentration of ships is not typical. Anyway, detecting maritime vessels in commercial ports can be an excellent indicator of economic activity, as maritime transport carries out more than 90\% of the world's trade\footnote{https://business.un.org/en/entities/13}.

In Figure~\ref{fig:covid-scenarios-a}(b), we show a campaign hospital being built in a field 31 miles outside of Moscow, Russia, to treat COVID-19 patients, as reported by Dave Mosher from Business Insider\footnote{https://www.businessinsider.com/coronavirus-covid-19-russian-hospital-field-near-moscow-satellite-photos-2020-3}. We were able to detect different construction-related equipment, such as trucks, tents, and excavators. Although, in this case, the location must be determined beforehand, our framework can keep track of the construction site proportions, which in turn can indicate the magnitude of the outbreak in that location. 
In Figure~\ref{fig:covid-scenarios-a}(c), we show a drive-through COVID-19 testing site built in Munich, Germany. As can be seen, our automated count is very close to the real number (over 60 vehicles\footnote{https://www.gim-international.com/content/news/satellite-imagery-covid-19-testing-facilities-in-munich-germany}) and could help authorities to measure attendance over time in health facilities. 

\begin{figure*}[!htb]
   \centering
   \renewcommand*{\arraystretch}{0.7}
   \subfigure[Salt Lake City International Airport (USA) before COVID-19 outbreak]{
     \begin{tikzpicture}
        \draw(0.0,0.0) node[inner sep=0pt] (img1) {
     \includegraphics[height=6.9cm,width=8.97cm]{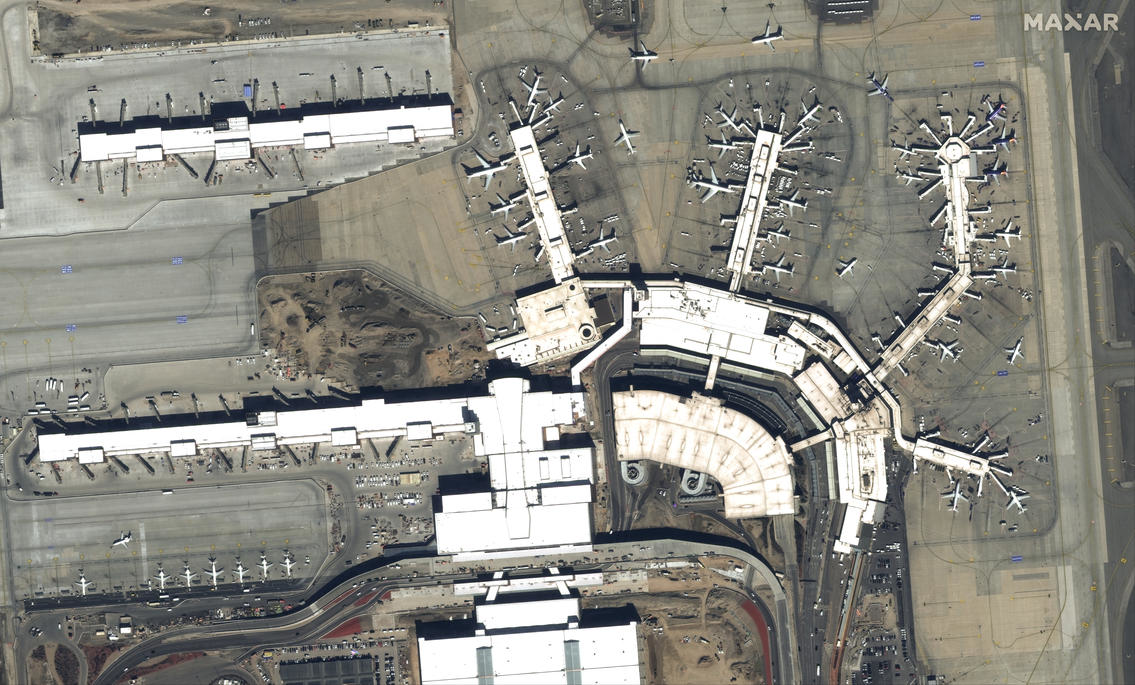}}; 
        \draw(9.15,0.0) node[inner sep=0pt] (img1) {
     \includegraphics[height=6.9cm,width=8.97cm]{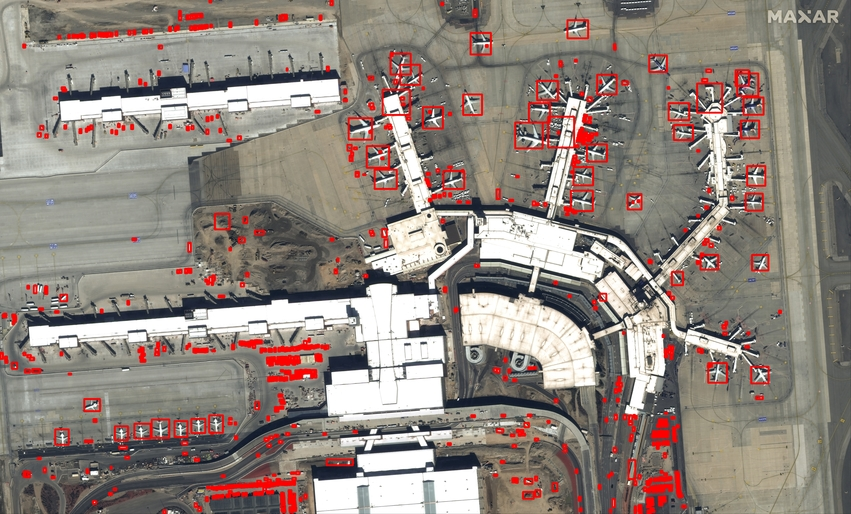}}; 
       \path[->](3.16,-2.75) node[black,very thick,fill=white,fill opacity=0.98, inner sep=0pt]  {\begin{tabular}{l}\textbf{\scriptsize Ground truth:}\\\scriptsize small-car = 664 \\\scriptsize plane = 38   \\\scriptsize \textbf{...}\end{tabular}};
       \path[->](12.25,-2.75) node[black,very thick,fill=white,fill opacity=0.98, inner sep=0pt]  {\begin{tabular}{l}\textbf{\scriptsize Detection result:}\\\scriptsize small-car = 665 \\\scriptsize plane = 42  \\\scriptsize \textbf{...}\end{tabular}};
     \end{tikzpicture}
   } \\
   \subfigure[Salt Lake City International Airport (USA) after COVID-19 outbreak]{
     \begin{tikzpicture}
        \draw(0.0,0.0) node[inner sep=0pt] (img1) {
     \includegraphics[height=6.9cm,width=8.97cm]{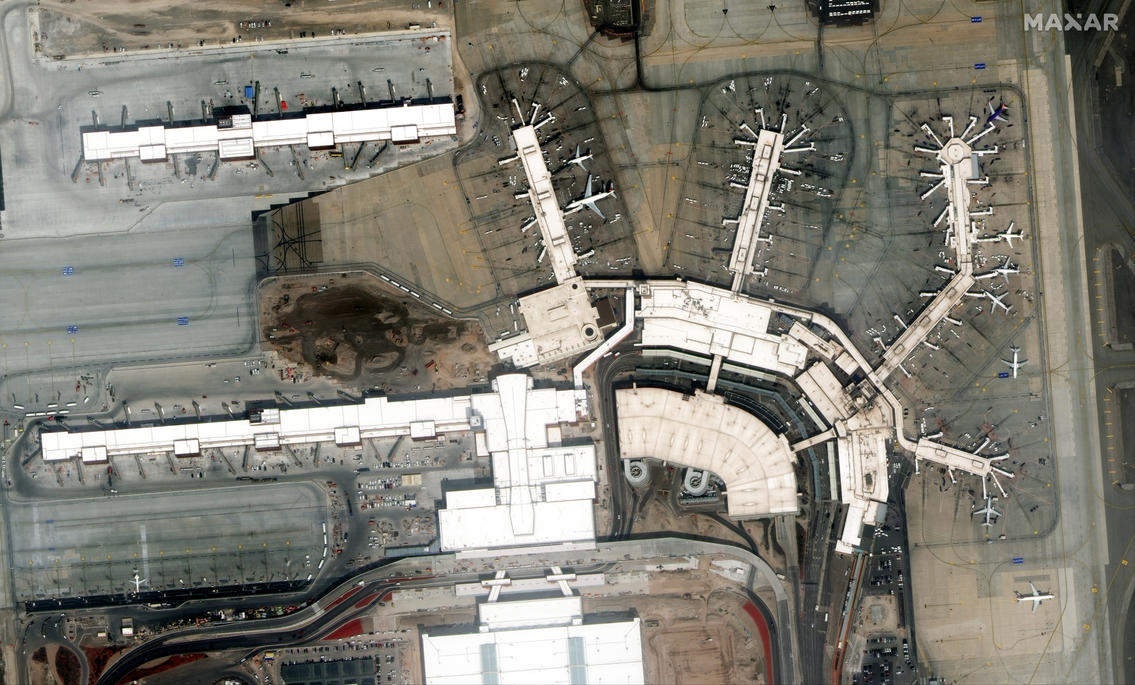}}; 
        \draw(9.15,0.0) node[inner sep=0pt] (img1) {
     \includegraphics[height=6.9cm,width=8.97cm]{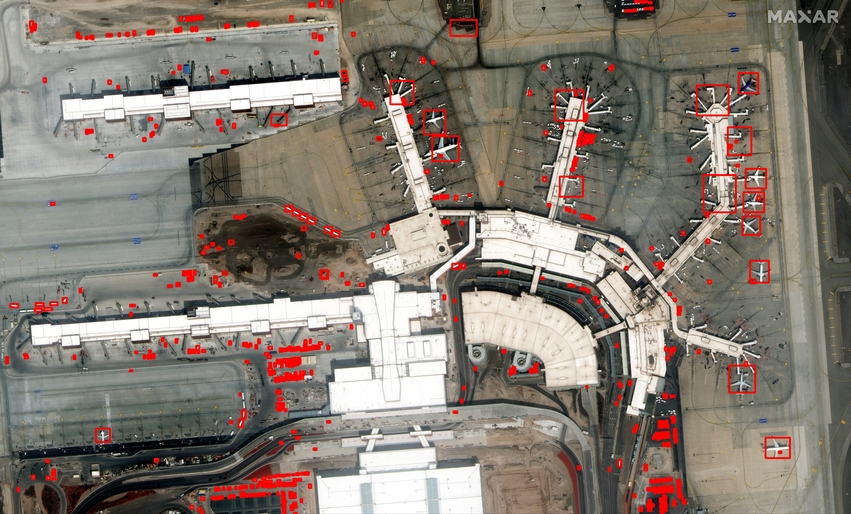}}; 
       \path[->](3.16,-2.75) node[black,very thick,fill=white,fill opacity=0.98, inner sep=0pt]  {\begin{tabular}{l}\textbf{\scriptsize Ground truth:}\\\scriptsize small-car = 570 \\\scriptsize plane = 10   \\\scriptsize \textbf{...}\end{tabular}};
       \path[->](12.25,-2.75) node[black,very thick,fill=white,fill opacity=0.98, inner sep=0pt]  {\begin{tabular}{l}\textbf{\scriptsize Detection result:}\\\scriptsize small-car = 528 \\\scriptsize plane = 16   \\\scriptsize \textbf{...}\end{tabular}};
     \end{tikzpicture}
   }
   \caption{Satellite imagery from world scenes before and after the COVID-19 outbreak and statistics about vehicles/infrastructure available. \textbf{Photo credit}: \textsc{Satellite image 2020 Maxar Technologies}.}
   \label{fig:covid-scenarios-b2} 
\end{figure*}

\begin{figure*}[!htb]
   \centering
   \renewcommand*{\arraystretch}{0.7}
   \subfigure[Golf course and a supermarket in Colorado (USA) before COVID-19 outbreak]{
     \begin{tikzpicture}
        \draw(0.0,0.0) node[inner sep=0pt] (img1) {
     \includegraphics[height=6.9cm,width=8.97cm]{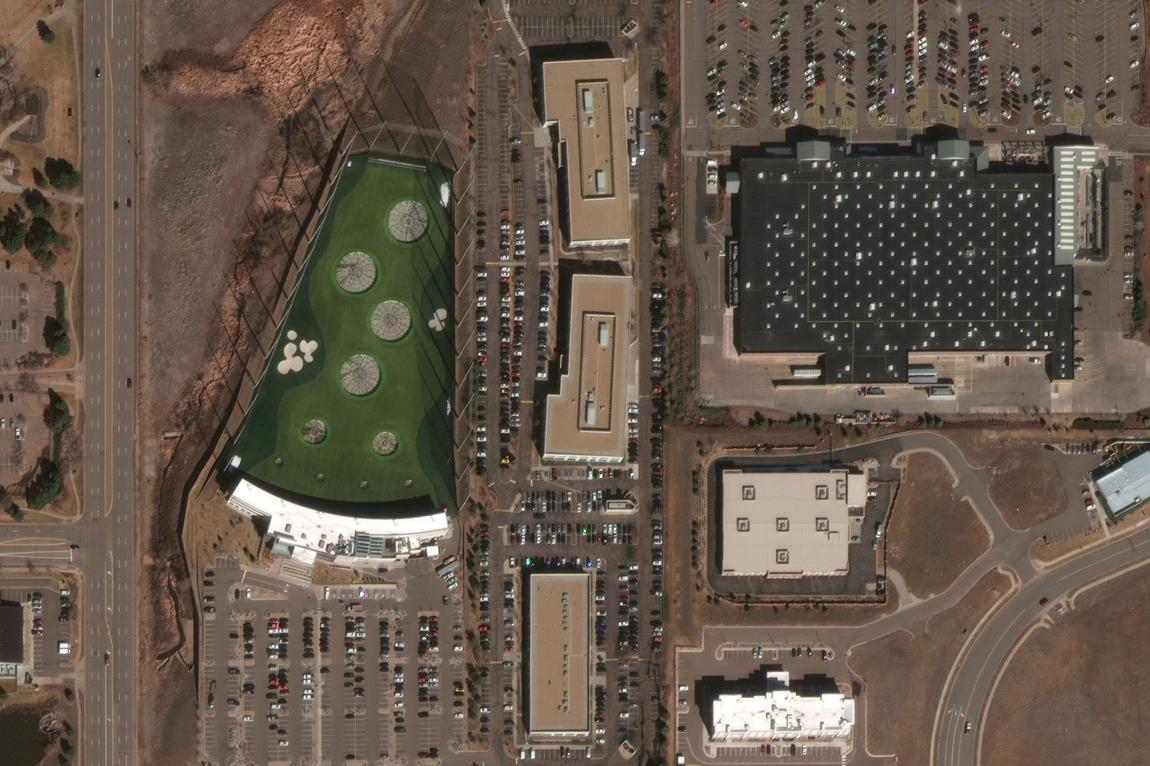}}; 
        \draw(9.15,0.0) node[inner sep=0pt] (img2) {
     \includegraphics[height=6.9cm,width=8.97cm]{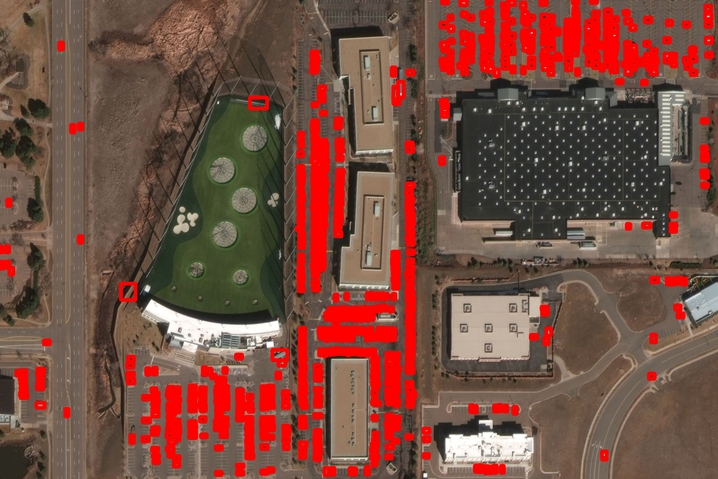}}; 
       \path[->](img1.north west) node[anchor=north west,black,very thick,fill=white,fill opacity=0.98, inner sep=0pt]  {\begin{tabular}{l}\textbf{\scriptsize Ground truth (golf-course):}\\\scriptsize small-car = 610\\\scriptsize \textbf{...}\end{tabular}};
      \path[->](img1.south east) node[anchor=south east,black,very thick,fill=white,fill opacity=0.98, inner sep=0pt]  {\begin{tabular}{l}\textbf{\scriptsize Ground truth (supermarket):}\\\scriptsize small-car = 292\\\scriptsize \textbf{...}\end{tabular}};
     
     \path[->](img2.north west) node[anchor=north west,black,very thick,fill=white,fill opacity=0.98, inner sep=0pt]  {\begin{tabular}{l}\textbf{\scriptsize Detection result (golf-course):}\\\scriptsize small-car = 631\\\scriptsize \textbf{...}\end{tabular}};
       
      \path[->](img2.south east) node[anchor=south east,black,very thick,fill=white,fill opacity=0.98, inner sep=0pt]  {\begin{tabular}{l}\textbf{\scriptsize Detection result (supermarket):}\\\scriptsize small-car = 297\\\scriptsize \textbf{...}\end{tabular}}; 
     
     \end{tikzpicture}
   } \\
   \subfigure[Golf course and a supermarket in Colorado (USA) after COVID-19 outbreak]{
     \begin{tikzpicture}
        \draw(0.0,0.0) node[inner sep=0pt] (img1) {
     \includegraphics[height=6.9cm,width=8.97cm]{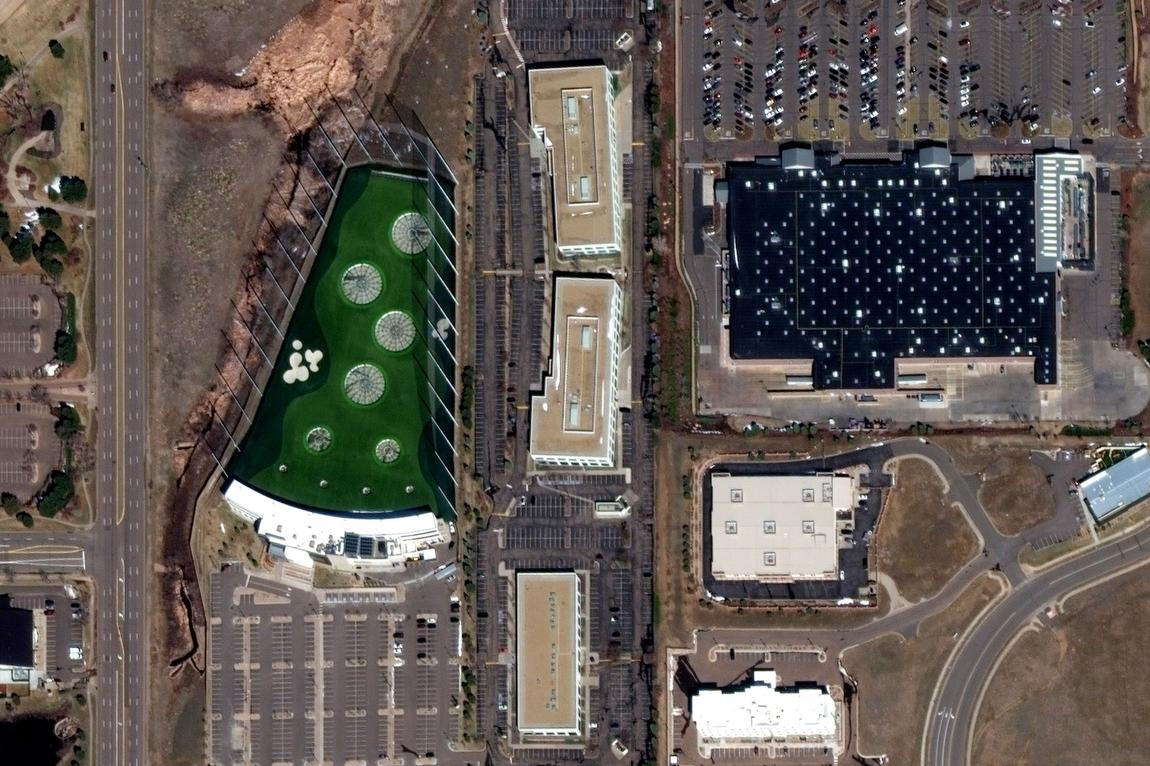}}; 
        \draw(9.15,0.0) node[inner sep=0pt] (img2) {
     \includegraphics[height=6.9cm,width=8.97cm]{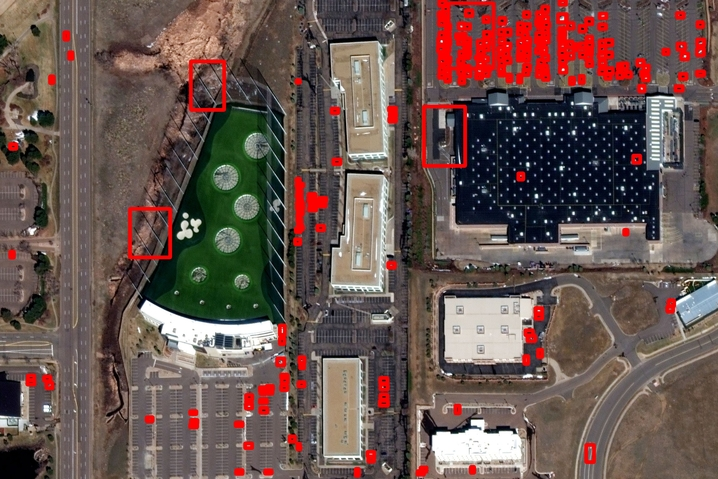}}; 
     \path[->](img1.north west) node[anchor=north west,black,very thick,fill=white,fill opacity=0.98, inner sep=0pt]  {\begin{tabular}{l}\textbf{\scriptsize Ground truth (golf-course):}\\\scriptsize small-car = 47\\\scriptsize \textbf{...}\end{tabular}};
       
      \path[->](img1.south east) node[anchor=south east,black,very thick,fill=white,fill opacity=0.98, inner sep=0pt]  {\begin{tabular}{l}\textbf{\scriptsize Ground truth (supermarket):}\\\scriptsize small-car = 219\\\scriptsize \textbf{...}\end{tabular}};
      
      \path[->](img2.north west) node[anchor=north west,black,very thick,fill=white,fill opacity=0.98, inner sep=0pt]  {\begin{tabular}{l}\textbf{\scriptsize Detection result (golf-course):}\\\scriptsize small-car = 76\\\scriptsize \textbf{...}\end{tabular}};
       
      \path[->](img2.south east) node[anchor=south east,black,very thick,fill=white,fill opacity=0.98, inner sep=0pt]  {\begin{tabular}{l}\textbf{\scriptsize Detection result (supermarket):}\\\scriptsize small-car = 174\\\scriptsize \textbf{...}\end{tabular}}; 
     \end{tikzpicture}
   }
   \caption{Satellite imagery from world scenes before and after the COVID-19 outbreak and statistics about vehicles/infrastructure available. \textbf{Photo credit}: \textsc{Satellite image 2020 Maxar Technologies}.}
   \label{fig:covid-scenarios-b3} 
\end{figure*}

\begin{figure*}[!htb]
   \centering
   \renewcommand*{\arraystretch}{0.7}
   \subfigure[Wuhan (China) toll plaza before COVID-19 outbreak]{
     \begin{tikzpicture}
        \draw(0.0,0.0) node[inner sep=0pt] (img1) {
     \includegraphics[height=4.5cm,width=8.97cm]{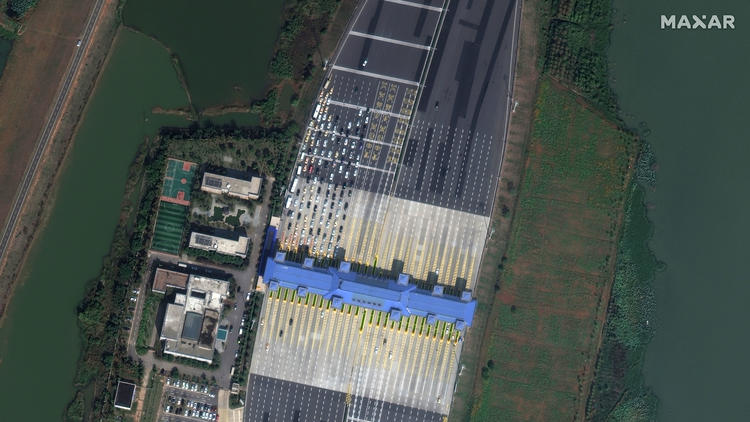}}; 
        \draw(9.15,0.0) node[inner sep=0pt] (img1) {
     \includegraphics[height=4.5cm,width=8.97cm]{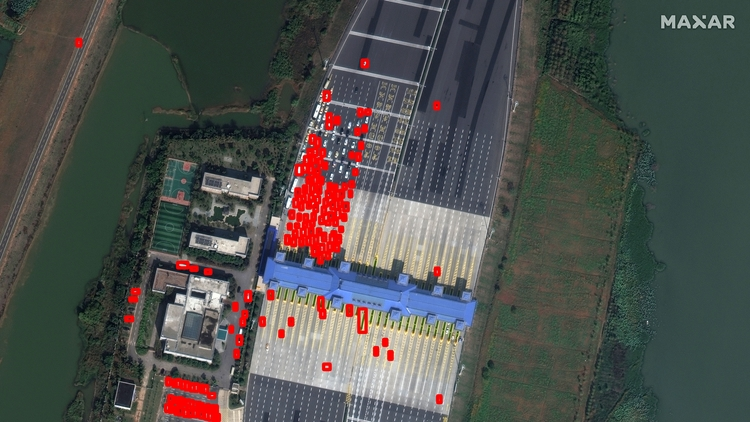}}; 
        \path[->](3.36,-1.75) node[black,very thick,fill=white,fill opacity=0.98, inner sep=0pt]  {\begin{tabular}{l}\textbf{\scriptsize Ground truth:}\\\scriptsize small-car = 201 \\\scriptsize \textbf{...}\end{tabular}};
       \path[->](12.40,-1.75) node[black,very thick,fill=white,fill opacity=0.98, inner sep=0pt]  {\begin{tabular}{l}\textbf{\scriptsize Detection result:}\\\scriptsize small-car = 139 \\\scriptsize \textbf{...}\end{tabular}};
     \end{tikzpicture}
   } \\
   \subfigure[Wuhan (China) toll plaza after COVID-19 outbreak]{
     \begin{tikzpicture}
        \draw(0.0,0.0) node[inner sep=0pt] (img1) {
     \includegraphics[height=4.5cm,width=8.97cm]{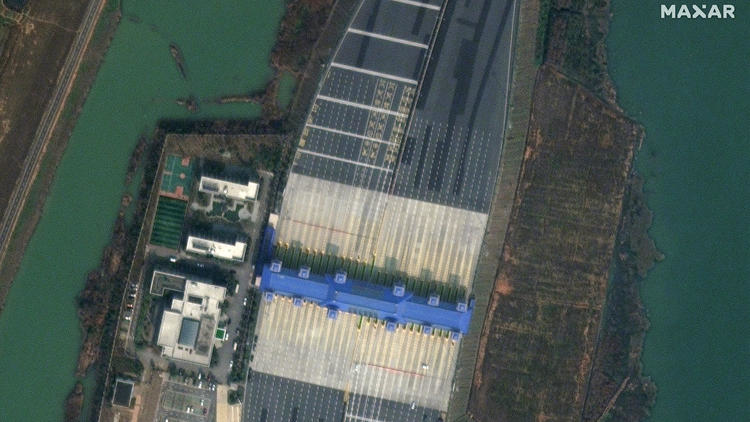}}; 
        \draw(9.15,0.0) node[inner sep=0pt] (img1) {
     \includegraphics[height=4.5cm,width=8.97cm]{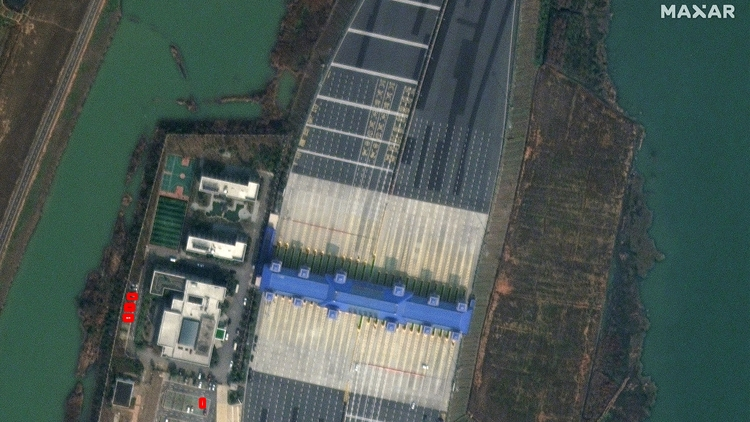}}; 
       \path[->](3.36,-1.75) node[black,very thick,fill=white,fill opacity=0.98, inner sep=0pt]  {\begin{tabular}{l}\textbf{\scriptsize Ground truth:}\\\scriptsize small-car = 33 \\\scriptsize \textbf{...}\end{tabular}};
       \path[->](12.40,-1.75) node[black,very thick,fill=white,fill opacity=0.98, inner sep=0pt]  {\begin{tabular}{l}\textbf{\scriptsize Detection result:}\\\scriptsize small-car = 4 \\\scriptsize \textbf{...}\end{tabular}};
     \end{tikzpicture}
   }
   \caption{Satellite imagery from world scenes before and after the COVID-19 outbreak and statistics about vehicles/infrastructure available. \textbf{Photo credit}: \textsc{Satellite image 2020 Maxar Technologies}.}
   \label{fig:covid-scenarios-b4} 
\end{figure*}

For some rare scenes, we were able to find the images before and after the pandemic. Thus, it was possible to use the temporal information to illustrate the economic effect of the outbreak. For example, Figure~\ref{fig:covid-scenarios-b1} show a car rental parking lot in the Phoenix Airport, Arizona, before and after the COVID-19 outbreak, respectively. As fewer people are traveling, the volume of parked vehicles increases substantially. This information serves not only to verify stay-at-home compliance but also to estimate the economic impact in a chain of companies such as car rental companies, airlines, insurance, etc. 

Figure~\ref{fig:covid-scenarios-b2} presents another example of a plausible economic indicator. It shows the number of planes in activity at the Salt Lake City International Airport (USA) before and after the COVID-19 outbreak, indicating that the pandemic has dramatically impacted the airplane travel segment. Our framework can automatically detect this decrease.

Figure~\ref{fig:covid-scenarios-b3} shows an interesting aspect. Before the pandemic it is possible to see a golf course (left) and a supermarket (top right) and their respective parking lots. The same location after the pandemic shows that the golf course parking lot is almost empty while the supermarket parking lot had a much smaller change in the number of cars. This information could be used to indicate the compliance to stay-at-home orders and the occurrence of panic buying.

Finally, Figure~\ref{fig:covid-scenarios-b4} shows car lines in a tollbooth at Wuhan, China, before the COVID-19 outbreak, and the same tollbooth empty after. This example shows how a simple redefinition of the list of strategic locations ({\it e.g.}, in this case, to tollbooths, highways, and border crossings) allows our framework to detect traffic jams and migration flows.

We annotated more than 16,000 vehicles on the images shown in Figures~\ref{fig:covid-scenarios-b1},~\ref{fig:covid-scenarios-b2},~\ref{fig:covid-scenarios-b3},~and~\ref{fig:covid-scenarios-b4} for performance analysis purposes. The detector achieved a $0.94$ mAP for airplanes, $0.66$ mAP for small cars, and a 10\% MAPE in object counting for images with more than 100 annotations. These results were slightly better than the ones obtained for the fMoW images in Section~\ref{sec:exp_temporal}. This outcome, allied to the fact that we were able to automatically perceive the impact of COVID-19 in real scenes ({\it e.g.}, decrease in aircraft on airport gates, increase in vehicles available for rental, small changes in parking occupancy of grocery stores), reinforces the proposed framework's aptness to the intended application.

\section{Discussion}~\label{sec:discussion}
We can automatically detect different types of vehicle from satellite imagery, which can be indicators for underlying economic activity. We describe our workflow in the context of enforcing a stay-at-home order (Section~\ref{sec:framework}), but it can be used for different purposes, as we discussed in Section~\ref{sec:casestudies}. By using two publicly available datasets, xView and fMoW, we estimated how well our detector worked for 60 distinct classes of objects (Section~\ref{sec:exp_detector}) and exemplified how well we can perceive the flow of these objects over time (Section~\ref{sec:exp_temporal}).

Adaptations of the proposed framework could measure many other indicators. For instance, counting trucks on roads and highways, trains on railways and stations, or containers in dry ports are all possibilities of economic indicators that can also point out problems in the supply chain. Observing tractors in rural areas can help evaluate agriculture activity, the same way the volume of vehicles in factories can assess industrial activity. The pool of indicators varies according to the local interests of each city. At this point, we can identify changes in the volume of objects using a small set of images and possibly classify these changes in different orders of magnitude. This information by itself could compose a dashboard to help decision-makers in spotting situations that demand immediate attention.

While human activities can be categorized in qualitative terms ({\it e.g.}, low, normal, high), economic activities are better represented as continuous values. However, any attempt to model and forecast changes in consolidated indicators (e.g., tourism revenue, local businesses' earnings, real estate vacancy rate) requires longer satellite image sequences with a high revisit rate and a history of indicator values. We are currently looking for a high-resolution collection of satellite images over time to evaluate our framework under real, continual circumstances. To this end, we are seeking partnerships with the industry and government agencies to gain access to such data. With that, hopefully, we will be able to provide a flexible tool that can be explored by authorities during this COVID-19 outbreak or in future events demanding a similar awareness over human activities.

It is worth noting that a framework based on satellite images has different limitations. The first one relates to the high cost of the data, especially when high-resolution images and high-frequency acquisition are required. Second, satellites hardly capture object dynamics within a day and cannot detect objects at night. Moreover, object visibility is affected by different factors, such as weather conditions (cloud cover) and satellite point of view (occlusions caused by trees and buildings). Finally, vehicle-based indicators cannot use ROIs with enclosed parking garages. Depending on the application's accuracy requirements, it may be necessary to combine satellite data with other data sources, such as mobile phone data, GPS signals, tollbooth records, etc. On the other hand, these alternative sources also have their limitations. For instance, smartphone data may not cover all models, operating systems, and carriers; GPS signals are affected by weather and buildings; most ROIs may not have a tollbooth nearby. Additionally, many developing nations may not have the infrastructure for aggregating these sources. Thus, in scenarios where a fast response is critical, satellite data stands out for its portability and coverage.

\ifCLASSOPTIONcompsoc
  \section*{Acknowledgments}
\else
  \section*{Acknowledgment}
\fi

Part of the equipment used in this project are supported by a grant (CNS-1513126) from the USA National Science Foundation. We gratefully acknowledge the support of NVIDIA Corporation with the donation of the Titan Xp GPU used for this research. Funding from the University of South Florida for the Institute for Artificial Intelligence (AI+X) is also acknowledged. The authors would like to thank also the research Brazilian agencies CNPq, CAPES and FAPESP. 

\ifCLASSOPTIONcaptionsoff
  \newpage
\fi

\bibliographystyle{IEEEtran}
\bibliography{bibliography}

\begin{thebibliography}{10}
\providecommand{\url}[1]{#1}
\csname url@samestyle\endcsname
\providecommand{\newblock}{\relax}
\providecommand{\bibinfo}[2]{#2}
\providecommand{\BIBentrySTDinterwordspacing}{\spaceskip=0pt\relax}
\providecommand{\BIBentryALTinterwordstretchfactor}{4}
\providecommand{\BIBentryALTinterwordspacing}{\spaceskip=\fontdimen2\font plus
\BIBentryALTinterwordstretchfactor\fontdimen3\font minus
  \fontdimen4\font\relax}
\providecommand{\BIBforeignlanguage}[2]{{%
\expandafter\ifx\csname l@#1\endcsname\relax
\typeout{** WARNING: IEEEtran.bst: No hyphenation pattern has been}%
\typeout{** loaded for the language `#1'. Using the pattern for}%
\typeout{** the default language instead.}%
\else
\language=\csname l@#1\endcsname
\fi
#2}}
\providecommand{\BIBdecl}{\relax}
\BIBdecl

\bibitem{ASADZADEH2016162}
S.~Asadzadeh and C.~R.~S. Filho, ``Investigating the capability of worldview-3
  superspectral data for direct hydrocarbon detection,'' \emph{Remote Sensing
  of Environment}, vol. 173, pp. 162--173, 2016.

\bibitem{xview2018}
D.~Lam, R.~Kuzma, K.~McGee, S.~Dooley, M.~Laielli, M.~Klaric, Y.~Bulatov, and
  B.~McCord, ``x{V}iew: {O}bjects in {C}ontext in {O}verhead {I}magery,''
  \emph{CoRR}, vol. abs/1802.07856, 2018.

\bibitem{christie2018functional}
G.~Christie, N.~Fendley, J.~Wilson, and R.~Mukherjee, ``Functional {M}ap of the
  {W}orld,'' in \emph{IEEE CVPR}, 2018, pp. 6172--6180.

\bibitem{DBLP:journals/corr/abs-1807-01232}
A.~V. Etten, D.~Lindenbaum, and T.~M. Bacastow, ``Spacenet: {A} remote sensing
  dataset and challenge series,'' \emph{CoRR}, vol. abs/1807.01232, 2018.

\bibitem{CreatingxBD}
R.~Gupta, B.~Goodman, N.~Patel, R.~Hosfelt, S.~Sajeev, E.~Heim, J.~Doshi,
  K.~Luscas, H.~Choset, and M.~Gaston, ``{Creating xBD}: A dataset for
  assessing building damage from satellite imagery,'' in \emph{IEEE CVPR},
  2019, pp. 10--17.

\bibitem{ROGERS1991}
D.~J. Rogers, ``Satellite imagery, tsetse and trypanosomiasis in {A}frica,''
  \emph{Preventive Veterinary Medicine}, vol.~11, no.~3, pp. 201--220, 1991.

\bibitem{ROGERS2002}
D.~J. Rogers, S.~E. Randolph, R.~W. Snow, and S.~I. Hay, ``Satellite imagery in
  the study and forecast of malaria,'' \emph{Nature}, vol. 415, no. 6872, pp.
  710--715, 2002.

\bibitem{Dister1997}
S.~W. Dister, D.~Fish, S.~M. Bros, D.~H. Frank, and B.~L. Wood, ``Landscape
  characterization of peridomestic risk for lyme disease using satellite
  imagery,'' \emph{The American Journal of Tropical Medicine and Hygiene},
  vol.~57, no.~6, pp. 687--692, 1997.

\bibitem{Cyranoski2009}
D.~Cyranoski, ``Putting china's wetlands on the map,'' \emph{Nature}, vol. 458,
  no. 134, 2009.

\bibitem{Ford2009}
T.~E. Ford, R.~R. Colwell, J.~B. Rose, S.~S. Morse, D.~J. Rogers, and T.~L.
  Yates, ``Using satellite images of environmental changes to predict
  infectious disease outbreaks,'' \emph{Emerging infectious diseases}, vol.~15,
  no.~9, pp. 1341--1346, 2009.

\bibitem{GARNI2014}
R.~Garni, A.~Tran, H.~Guis, T.~Baldet, K.~Benallal, S.~Boubidi, and Z.~Harrat,
  ``Remote sensing, land cover changes, and vector-borne diseases: Use of high
  spatial resolution satellite imagery to map the risk of occurrence of
  cutaneous leishmaniasis in ghardaia, algeria,'' \emph{Infection, Genetics and
  Evolution}, vol.~28, pp. 725--734, 2014.

\bibitem{Ghosh2010}
G.~Tilottama, R.~Powell, C.~Elvidge, K.~Baugh, and P.~Sutton, ``Shedding light
  on the global distribution of economic activity,'' \emph{The Open Geography
  Journal}, vol.~3, 01 2010.

\bibitem{Hu2019}
Y.~Hu and J.~Yao, ``Illuminating economic growth,'' \emph{IMF Working Papers},
  vol.~19, p.~1, 01 2019.

\bibitem{Elvidge2009}
C.~D. Elvidge, P.~C. Sutton, T.~Ghosh, B.~T. Tuttle, K.~E. Baugh, B.~Bhaduri,
  and E.~Bright, ``A global poverty map derived from satellite data,''
  \emph{Computers and Geosciences}, vol.~35, no.~8, pp. 1652--1660, 2009.

\bibitem{Noor2008}
A.~M. Noor, V.~A. Alegana, P.~W. Gething, A.~J. Tatem, and R.~W. Snow, ``Using
  remotely sensed night-time light as a proxy for poverty in africa,''
  \emph{Population Health Metrics}, vol.~6, no.~5, 2008.

\bibitem{Ghosh2010a}
G.~Tilottama, C.~Elvidge, P.~Sutton, K.~Baugh, and D.~Ziskin, ``Estimating the
  information and technology development index (idi) using nighttime satellite
  imagery,'' \emph{Proceedings of the Asia-Pacific Advanced Network}, vol.~30,
  2010.

\bibitem{Jean2016}
N.~Jean, M.~Burke, M.~Xie, W.~M. Davis, D.~B. Lobell, and S.~Ermon, ``Combining
  satellite imagery and machine learning to predict poverty,'' \emph{Science},
  vol. 353, no. 6301, pp. 790--794, 2016.

\bibitem{Suraj2017}
P.~K. Suraj, A.~Gupta, M.~Sharma, S.~B. Paul, and S.~Banerjee, ``On monitoring
  development indicators using high resolution satellite images,'' \emph{CoRR},
  vol. abs/1712.02282, 2017.

\bibitem{Lecun2015}
Y.~Lecun, Y.~Bengio, and G.~Hinton, ``Deep learning,'' \emph{Nature}, vol. 521,
  no. 7553, pp. 436--444, 5 2015.

\bibitem{Yeh2020}
C.~Yeh, A.~Perez, A.~Driscoll, G.~Azzari, Z.~Tang, D.~Lobell, S.~Ermon, and
  M.~Burke, ``Using publicly available satellite imagery and deep learning to
  understand economic well-being in africa,'' \emph{Nature Communications},
  vol.~11, no. 2583, 2020.

\bibitem{ganguli2019predicting}
S.~Ganguli, J.~Dunnmon, and D.~Hau, ``Predicting food security outcomes using
  convolutional neural networks (cnns) for satellite tasking,'' \emph{CoRR},
  vol. abs/1902.05433, 2019.

\bibitem{Fukushima1980}
K.~Fukushima, ``{N}eocognitron: {A} self-organizing neural network model for a
  mechanism of pattern recognition unaffected by shift in position,''
  \emph{Biological Cybernetics}, vol.~36, pp. 193--202, 1980.

\bibitem{Krizhevsky2012}
A.~Krizhevsky, I.~Sutskever, and G.~E. Hinton, ``Imagenet classification with
  deep convolutional neural networks,'' in \emph{Advances in Neural Information
  Processing Systems 25}, 2012, pp. 1097--1105.

\bibitem{Kampffmeyer_2016_CVPR_Workshops}
M.~Kampffmeyer, A.-B. Salberg, and R.~Jenssen, ``Semantic segmentation of small
  objects and modeling of uncertainty in urban remote sensing images using deep
  convolutional neural networks,'' in \emph{IEEE CVPR Workshops}, June 2016,
  pp. 1--9.

\bibitem{Guirado2019}
E.~Guirado, S.~Tabik, M.~Rivas, A.~Domingo, and F.~Herrera, ``Whale counting in
  satellite and aerial images with deep learning,'' \emph{Scientific Reports},
  vol.~9, no.~1, 2019.

\bibitem{8698456}
R.~{Minetto}, M.~{Pamplona Segundo}, and S.~{Sarkar}, ``Hydra: An ensemble of
  convolutional neural networks for geospatial land classification,''
  \emph{IEEE TGRS}, vol.~57, no.~9, pp. 6530--6541, 2019.

\bibitem{8713925}
F.~{Hu}, G.~{Xia}, W.~{Yang}, and L.~{Zhang}, ``Mining deep semantic
  representations for scene classification of high-resolution remote sensing
  imagery,'' \emph{IEEE Transactions on Big Data}, 2019.

\bibitem{8880494}
X.~{Tong}, G.~{Xia}, F.~{Hu}, Y.~{Zhong}, M.~{Datcu}, and L.~{Zhang},
  ``Exploiting deep features for remote sensing image retrieval: A systematic
  investigation,'' \emph{IEEE Transactions on Big Data}, 2019.

\bibitem{7937881}
S.~{Lei}, Z.~{Shi}, and Z.~{Zou}, ``Super-resolution for remote sensing images
  via local–global combined network,'' \emph{IEEE Geoscience and Remote
  Sensing Letters}, vol.~14, no.~8, pp. 1243--1247, 2017.

\bibitem{Kulp2019}
S.~A. Kulp and B.~H. Strauss, ``New elevation data triple estimates of global
  vulnerability to sea-level rise and coastal flooding,'' \emph{Nature
  communications}, vol.~10, no.~1, 2019.

\bibitem{8316243}
Q.~{Zhang}, Q.~{Yuan}, C.~{Zeng}, X.~{Li}, and Y.~{Wei}, ``Missing data
  reconstruction in remote sensing image with a unified
  spatial–temporal–spectral deep convolutional neural network,'' \emph{IEEE
  TGRS}, vol.~56, no.~8, pp. 4274--4288, 2018.

\bibitem{8707405}
G.~{Rotich}, S.~{Aakur}, R.~{Minetto}, M.~P. {Segundo}, and S.~{Sarkar},
  ``Using semantic relationships among objects for geospatial land use
  classification,'' in \emph{IEEE AIPR Workshop}, 2018, pp. 1--7.

\bibitem{Leotta_2019_CVPR_Workshops}
M.~J. Leotta, C.~Long, B.~Jacquet, M.~Zins, D.~Lipsa, J.~Shan, B.~Xu, Z.~Li,
  X.~Zhang, S.-F. Chang, M.~Purri, J.~Xue, and K.~Dana, ``Urban semantic 3d
  reconstruction from multiview satellite imagery,'' in \emph{IEEE CVPR
  Workshops}, June 2019, pp. 1--10.

\bibitem{8241773}
Y.~{Zhan}, D.~{Hu}, Y.~{Wang}, and X.~{Yu}, ``Semisupervised hyperspectral
  image classification based on generative adversarial networks,'' \emph{IEEE
  GRSL}, vol.~15, no.~2, pp. 212--216, 2018.

\bibitem{liu2016ssd}
W.~Liu, D.~Anguelov, D.~Erhan, C.~Szegedy, S.~Reed, C.-Y. Fu, and A.~C. Berg,
  ``{SSD}: {S}ingle {S}hot {M}ultibox {D}etector,'' in \emph{ECCV}, 2016, pp.
  21--37.

\bibitem{Felzenszwalb2010}
P.~F. Felzenszwalb, R.~B. Girshick, D.~McAllester, and D.~Ramanan, ``Object
  detection with discriminatively trained part-based models,'' \emph{IEEE
  TPAMI}, vol.~32, no.~9, pp. 1627--1645, Sep. 2010.

\bibitem{lin2020}
T.~{Lin}, P.~{Goyal}, R.~{Girshick}, K.~{He}, and P.~{Dollár}, ``Focal loss
  for dense object detection,'' \emph{IEEE TPAMI}, vol.~42, no.~2, pp.
  318--327, 2020.

\bibitem{Zhou2018}
P.~{Zhou}, B.~{Ni}, C.~{Geng}, J.~{Hu}, and Y.~{Xu}, ``Scale-transferrable
  object detection,'' in \emph{IEEE CVPR}, 2018, pp. 528--537.

\bibitem{Hu2018}
H.~{Hu}, J.~{Gu}, Z.~{Zhang}, J.~{Dai}, and Y.~{Wei}, ``Relation networks for
  object detection,'' in \emph{IEEE CVPR}, 2018, pp. 3588--3597.

\bibitem{Shumway2005}
R.~H. Shumway and D.~S. Stoffer, \emph{Time Series Analysis and Its
  Applications}.\hskip 1em plus 0.5em minus 0.4em\relax Berlin, Heidelberg:
  Springer-Verlag, 2005.

\bibitem{Gama2014}
J.~a. Gama, I.~\v{Z}liobaitundefined, A.~Bifet, M.~Pechenizkiy, and
  A.~Bouchachia, ``A survey on concept drift adaptation,'' \emph{ACM Comput.
  Surv.}, vol.~46, no.~4, Mar. 2014.

\bibitem{Lai2018}
G.~Lai, W.-C. Chang, Y.~Yang, and H.~Liu, ``Modeling long- and short-term
  temporal patterns with deep neural networks,'' in \emph{International ACM
  SIGIR Conference on Research and Development in Information Retrieval}, 2018,
  pp. 95--104.

\bibitem{Qin2017}
Y.~Qin, D.~Song, H.~Cheng, W.~Cheng, G.~Jiang, and G.~W. Cottrell, ``A
  dual-stage attention-based recurrent neural network for time series
  prediction,'' in \emph{IJCAI}, 2017, pp. 2627--2633.

\bibitem{DBLP:journals/corr/HendersonF16}
P.~Henderson and V.~Ferrari, ``End-to-end training of object class detectors
  for mean average precision,'' \emph{CoRR}, vol. abs/1607.03476, 2016.

\bibitem{li2015}
H.~Li, L.~Jing, Y.~Tang, Q.~Liu, H.~Ding, Z.~Sun, and Y.~Chen, ``Assessment of
  pan-sharpening methods applied to worldview-2 image fusion,'' in \emph{IEEE
  IGARSS}, 2015, pp. 3302--3305.

\bibitem{zsolt2018}
Z.~Katona, M.~Painter, P.~N. Patatoukas, and J.~Zeng, ``On the capital market
  consequences of alternative data: Evidence from outer space,'' in \emph{9th
  Miami Behavioral Finance Conference}, 2018.

\end{thebibliography}

\vspace{-20pt}

\begin{IEEEbiography} [{\includegraphics[width=1in,height=1.25in,clip,keepaspectratio]{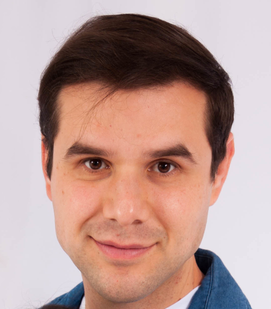}}]{Rodrigo Minetto} is an assistant  professor  at  Federal  University  of  Technology - Paran\'{a} (UTFPR) - Brazil. He received the Ph.D. in computer science in 2012 from University of Campinas (UNICAMP), Brazil and Universit\'{e} Pierre et Marie Curie, France (UPMC).  His research interests include image processing, computer vision and machine learning. Currently he is a visiting scholar at University of South Florida (USF), USA.\end{IEEEbiography}

\begin{IEEEbiography}[{\includegraphics[width=1in,height=1.25in,clip,keepaspectratio]{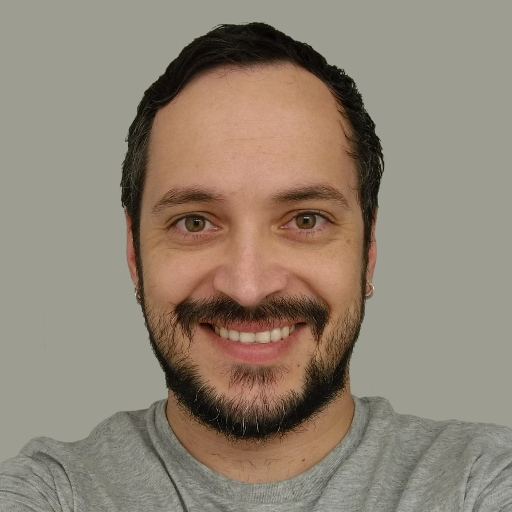}}]{Maur\'icio Pamplona Segundo} is a Postdoctoral Researcher at the Institute for Artificial Intelligence (AI+X), University of South Florida. He received his BSc, MSc and DSc in Computer Science from the Federal University of Paran\'{a} (UFPR). His areas of expertise are computer vision and pattern recognition, and his research interests include biometrics, remote sensing, 3D reconstruction, accessibility tools, and vision-based automation.\end{IEEEbiography}

\begin{IEEEbiography}[{\includegraphics[width=1in,height=1.25in,clip,keepaspectratio]{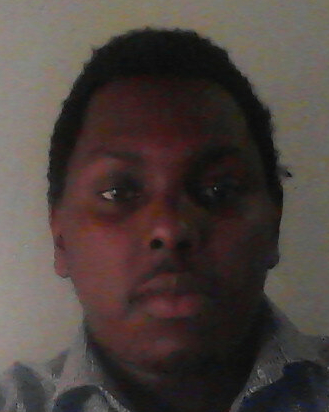}}]{Gilbert Rotich} is a Ph.D. candidate at University of South Florida. He received his BSc in Computer Engineering from Bethune-Cookman University . His research interests include computer vision, machine learning, deep learning and remote sensing. \end{IEEEbiography}

\begin{IEEEbiography}[{\includegraphics[width=1in,height=1.25in,clip,keepaspectratio]{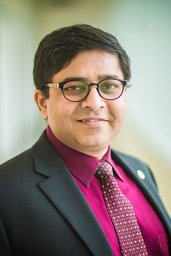}}]{Sudeep Sarkar} is a professor of Computer Science and Engineering and Associate Vice President for Research \& Innovation at the University of South Florida in Tampa. He received his MS and PhD degrees in Electrical Engineering, on a University Presidential Fellowship, from The Ohio State University. He is the recipient of the National Science Foundation CAREER award in 1994, the USF Teaching Incentive Program Award for Undergraduate Teaching Excellence in 1997, the Outstanding Undergraduate Teaching Award in 1998, and the Theodore and Venette Askounes-Ashford Distinguished Scholar Award in 2004. He is a Fellow of the American Association for the Advancement of Science (AAAS), Institute of Electrical and Electronics Engineers (IEEE), American Institute for Medical and Biological Engineering (AIMBE), and International Association for Pattern Recognition (IAPR); and a charter member and member of the Board of Directors of the National Academy of Inventors (NAI). He has 25 year expertise in computer vision and pattern recognition algorithms and systems, holds three U.S. patents and has published high-impact journal and conference papers.\end{IEEEbiography}

\end{document}